\documentclass[10pt,journal,compsoc]{IEEEtran}
\usepackage[cmex10]{amsmath}
\usepackage{algorithmic}
\usepackage{algorithm}
\usepackage{array}
\usepackage{textcomp}
\usepackage{stfloats}
\usepackage{url}
\usepackage{verbatim}
\usepackage{graphicx}
\usepackage{hyperref}
\usepackage{color}
\usepackage{xcolor}
\usepackage{bbm}
\usepackage{booktabs}
\usepackage{multirow}
\usepackage{bm}
\usepackage{colortbl}
\usepackage{amssymb}
\usepackage{enumitem}
\usepackage{rotating}
\usepackage[section]{placeins}
\usepackage{cleveref}
\crefname{figure}{Figure}{Figures}  
\crefname{table}{Table}{Tables} 
\crefname{equation}{Eq.}{Eqs.}
\crefname{section}{Section}{Section}

\ifCLASSOPTIONcompsoc
  \usepackage[nocompress]{cite}
\else
  \usepackage{cite}
\fi
\ifCLASSINFOpdf
\else
\fi
\ifCLASSOPTIONcompsoc
 \usepackage[caption=false,font=footnotesize,labelfont=sf,textfont=sf]{subfig}
\else
 \usepackage[caption=false,font=footnotesize]{subfig}
\fi

\begin{document}
%
\title{Learning ORDER-Aware Multimodal Representations for Composite Materials Design}
%
%
%
%

\author{Xinyao Li,
        Hangwei Qian,
        Jingjing Li,
        Lei Zhu,
        Ivor Tsang
\IEEEcompsocitemizethanks{
\IEEEcompsocthanksitem Work done during Xinyao Li's internship at A*STAR CFAR, Singapore.
\IEEEcompsocthanksitem Xinyao Li, Jingjing Li are with University of Electronic Science and Technology of China, Chengdu 610054, China.
\IEEEcompsocthanksitem Lei Zhu is with Tongji University, China.
\IEEEcompsocthanksitem Hangwei Qian, Ivor Tsang are with A*STAR CFAR, Singapore.
}
\thanks{Manuscript received ; revised  .}}

\markboth{under review}%
{Xinyao Li \MakeLowercase{\textit{et al.}}: xx}

%


\IEEEtitleabstractindextext{%
\begin{abstract}
Artificial intelligence (AI) has shown remarkable success in materials discovery and property prediction, particularly for crystalline and polymer systems where material properties and structures are dominated by discrete graph representations. Such graph-central paradigm breaks down on composite materials, which possess continuous and nonlinear design spaces.  General composite descriptors, e.g., fiber volume and misalignment angle, cannot fully capture the fiber distributions that determine microstructural characteristics, necessitating the integration of heterogeneous data sources through multimodal learning. Existing alignment-oriented multimodal frameworks have proven effective on abundant crystal or polymer data under discrete, unique graph-property mapping assumptions, but fail to address the highly continuous composite design space under extreme data scarcity. In this work, we introduce ORDinal-aware imagE-tabulaR alignment (ORDER), a multimodal pretraining framework that establishes \textbf{ordinality} as a core principle for composite material representations. ORDER ensures that materials with similar target properties occupy nearby regions in the latent space, which effectively preserves the continuous nature of composite properties and enables meaningful interpolation between sparsely observed designs. We evaluate ORDER on a public Nanofiber-reinforced composite dataset and an internally curated carbon fiber T700 dataset. ORDER and its variants consistently outperform both alignment-oriented and customized property-aware contrastive baselines across property prediction, cross-modal retrieval, and microstructure generation tasks. We further introduce physics-based ordinal surrogate signals avoiding the need for full property annotation during pretrain. Our work demonstrates learning semantically continuous multimodal features are fundamental for composite materials, and provides a reliable pathway toward data-efficient universal multimodal intelligent systems.
\end{abstract}
}

\maketitle

\IEEEdisplaynontitleabstractindextext

%
\IEEEpeerreviewmaketitle

\section{Introduction}\label{sec:introduction}

Materials science forms the foundation for technological innovation across diverse fields, from semiconductors and catalysis to energy storage and biomedicine~\cite{pyzer2022accelerating}. Historically, materials discovery has relied on heuristic experimentation, theory, or computational approaches~\cite{himanen2019data,schleder2019dft}. While foundational, these methods remain labor-intensive and time-consuming~\cite{greenaway2021integrating,talapatra2019experiment},  requiring years to progress from discovery to deployment. Recent advances in machine learning, especially deep learning, have inspired data-driven techniques in materials science~\cite{wei2019machine,decost2020scientific}, substantially accelerating the research process~\cite{horton2025accelerated}.
These methods extract vast knowledge and patterns from large-scale materials databases. They primarily operate on string descriptors or graph structures to encode key information such as chemical compositions, crystal parameters, processing conditions, and have demonstrated remarkable effectiveness in tasks including property prediction~\cite{huang2025unified,moro2025multimodal}, materials design~\cite{pollice2021data}, and structure-application mapping~\cite{khan2025connecting}. While recent characterization techniques generate diverse data modalities, such as microscopy images revealing microstructural features~\cite{horie2025bidirectional} and X-ray diffraction patterns providing crystallographic information~\cite{oh2025taking}, they typically serve as complementary verification or alternative representations of information already encoded in material graph structures. For widely-researched materials like crystals and polymers, their graph representations sufficiently reflect material structure and properties in most applications \cite{xie2018crystal,aldeghi2022graph}.

However, this landscape changes for composite materials. Unlike crystals or polymers, composite properties are determined by the continuous, nonlinear, and infinitely variable fiber distributions within the polymer matrix, which discrete graph-based representations cannot encode.  General tabular descriptors for composites (e.g., fiber volume fraction and mean misalignment angle) constrain only high-level material compositions but fail to capture the spatial fiber distributions. Furthermore, the nonlinear property-structure relationships result in dramatic property changes triggered by minor changes in fiber arrangements \cite{makni2020synergistic}. Therefore, the fiber spatial distributions reflected in microscopy images become essential for composite materials, motivating us to unify coarse compositional control (in tabular descriptors) with actual fiber distributions (microstructural images) using multimodal pretraining techniques, such as  Contrastive Language-Image Pretraining (CLIP) \cite{radford2021learning}.
CLIP establishes a shared latent space where paired features (such as images and texts) from the same sample are drawn together while unpaired ones are pushed apart. This alignment framework has been extended to various modalities including tabular data~\cite{hager2023best,jiang2024tabular}, videos~\cite{xu2021videoclip}, and audio~\cite{guzhov2022audioclip}. It has also been proven effective in material science: MultiMat~\cite{moro2025multimodal} constructs a unified latent space for crystalline with modalities including density of states, charge density, and text descriptions, enabling materials property prediction and  discovery. Khan \textit{et al.}\cite{khan2025connecting} leverage powder X-ray diffraction patterns, chemical precursors, and crystal graphs to create synthesis-to-application maps for metal-organic frameworks. Huang \textit{et al.}\cite{huang2025unified} develop a multimodal dataset encompassing chemical precursors, 2D graphs, 3D geometries, fingerprints, and textual descriptions for polymer materials.

These approaches demonstrate strong performance on abundant crystal and polymer data, whose multimodal representation pairs, e.g., graph structures and target properties, are typically discrete and uniquely associated.  However, these conditions break down for composite materials, where fiber distribution, orientation, and density vary continuously within the design space. Moreover, the extreme data scarcity in composite materials poses a critical challenge: with limited multimodal samples (generally hundreds) representing less than 0.001\% of possible configurations, how to enable reliable materials design and understand across unobserved regions of this vast design space? Current multimodal methods for material science~\cite{sgpt,moro2025multimodal,khan2025connecting} treats materials as discrete entities to be matched across modalities, which lack the ability to interpolate between sparse samples or preserve the property continuity. Such continuity is physically grounded and crucial since composite mechanical properties, e.g., tensile strength and elongation vary smoothly with fiber volume fraction and misalignment angle.

In this work, we propose ORDinal-aware imagE-tabulaR alignment (ORDER), a multimodal pretraining framework that constructs a latent space reflecting both cross-modal alignment and ordinal property characteristics. Beyond mere alignment between tabular descriptors and microstructural images, we identify \emph{physical ordinality} as a crucial factor to consider when building the multimodal latent space: incremental increases in descriptors, e.g., fiber density corresponds to denser cross-sections in micrographs and  higher strength. ORDER ensures both the image and tabular embeddings of denser materials are consistently mapped further along the same direction in the latent space than sparser ones, thereby enabling reliable interpolation to unobserved design points from only hundreds of samples.  
Our key design to achieve this is ordinal-aware contrastive learning within each modality, which draws closer samples with similar target properties. To optimize against current multimodal contrastive learning process that equally separates any mismatch pair while not disturbing its cross-modal alignment effect, one straightforward solution is to weight the two objectives appropriately but requires grid-search on validation data. To achieve adaptive and dynamic weighting on any dataset, we further introduce preference-guided multitask learning~\cite{mahapatra2020multi} to ensure Pareto optimal of the bi-objective optimization process. Addressing annotation scarcity during pretraining, we additionally derive physics-based ordinal surrogate signals from constitutive material models as label-efficient proxies for ground-truth property labels, enabling effective pretraining when property measurements are unavailable. In accordance with prior work~\cite{sgpt}, we focus on aligning tabular and visual modalities, as descriptor and property information is typically documented in tabular form while microscopy images provide complementary microstructural details.
While existing approaches usually resort to training smaller neural networks~\cite{he2016deep} due to the limited multimodal composite datasets, we instead propose to adapt the vision transformer~\cite{dosovitskiy2020image} from the pretrained CLIP model~\cite{radford2021learning} using parameter-efficient fine-tuning (PEFT)~\cite{hu2022lora}.

ORDER aims to address a practical bottleneck in composite material design: characterizing materials with target mechanical properties demands expensive physical testing or high-fidelity simulation, while spatial microstructural detail remains invisible from coarse tabular descriptors alone. Applicable whether single-modality or multimodal data is available, ORDER aids materials design from various aspects, including but not limited to the following: 
\begin{itemize}[leftmargin=14pt,noitemsep,topsep=3pt]
  \item \textbf{Property prediction} estimates key mechanical properties such as tensile yield   strength and elongation from tabular descriptors and/or microstructural images, enabling faster investigation of material behavior while substantially reducing experimental burden and costs.
  \item \textbf{Cross-modal retrieval}. Given a set of target descriptor or an existing micrograph, ORDER rapidly retrieves the most mechanically and structurally similar material candidates from available database, supporting targeted exploration and experimental planning.
  \item \textbf{Microstructure generation} synthesizes realistic composite microstructures conditioned on tabular inputs, providing quick inspection of spatial fiber distributions (fiber count, density, misalignment angle, etc.) without the time-consuming and costly fabrication process.
\end{itemize}
These capabilities are grounded with various experiments in the paper. We evaluate ORDER on a public Nanofiber-reinforced composite dataset~\cite{sgpt} and our in-house CF-T700 Composite dataset, the first multimodal benchmark for this high-strength carbon fiber material. We further benchmark against two property-aware contrastive baselines, DWCL and Triplet by adapting established loss functions~\cite{khosla2020supervised,schroff2015facenet} to the continuous space. Experiments show ORDER variants achieve numerically superior performances and physically valid results, confirming its practical values.

\section{Results}

\subsection{Construction of Composite dataset}
\label{sec:dataset}
We propose the multimodal CFRP dataset (referred to as the Composite data for the rest of the paper) to enable the construction of vision-tabular models. CFRP is a composite material made of a polymer matrix (e.g., epoxy) and	carbon fibers (as reinforcement). Specifically, we focus on the varying random unidirectional carbon fibers in epoxy matrix of the CF-T700 material. We alter the volume fraction (Vf) in the composite and explore different mean misalignment angle (MMA) of the fibers as descriptors. Based on the Representative Volume Element (RVE) \cite{bargmann2018generation} method  from the Ansys Material Designer software, we simulate the corresponding microstructures and target properties (tensile yield strength, elongation) of the input descriptors. We modeled 436 different descriptor combinations with Vf ranging from 0.32 to 0.65 and MMA ranging from 0 to 5 degrees, and obtain their corresponding tabular-image pairs for our Composite dataset. More details are in \cref{sec:method_data}.

\subsection{Ordinal-aware image-tabular alignment}
\label{sec:order}

\begin{figure*}[!t]
  \centering
  \includegraphics[width=0.89\textwidth]{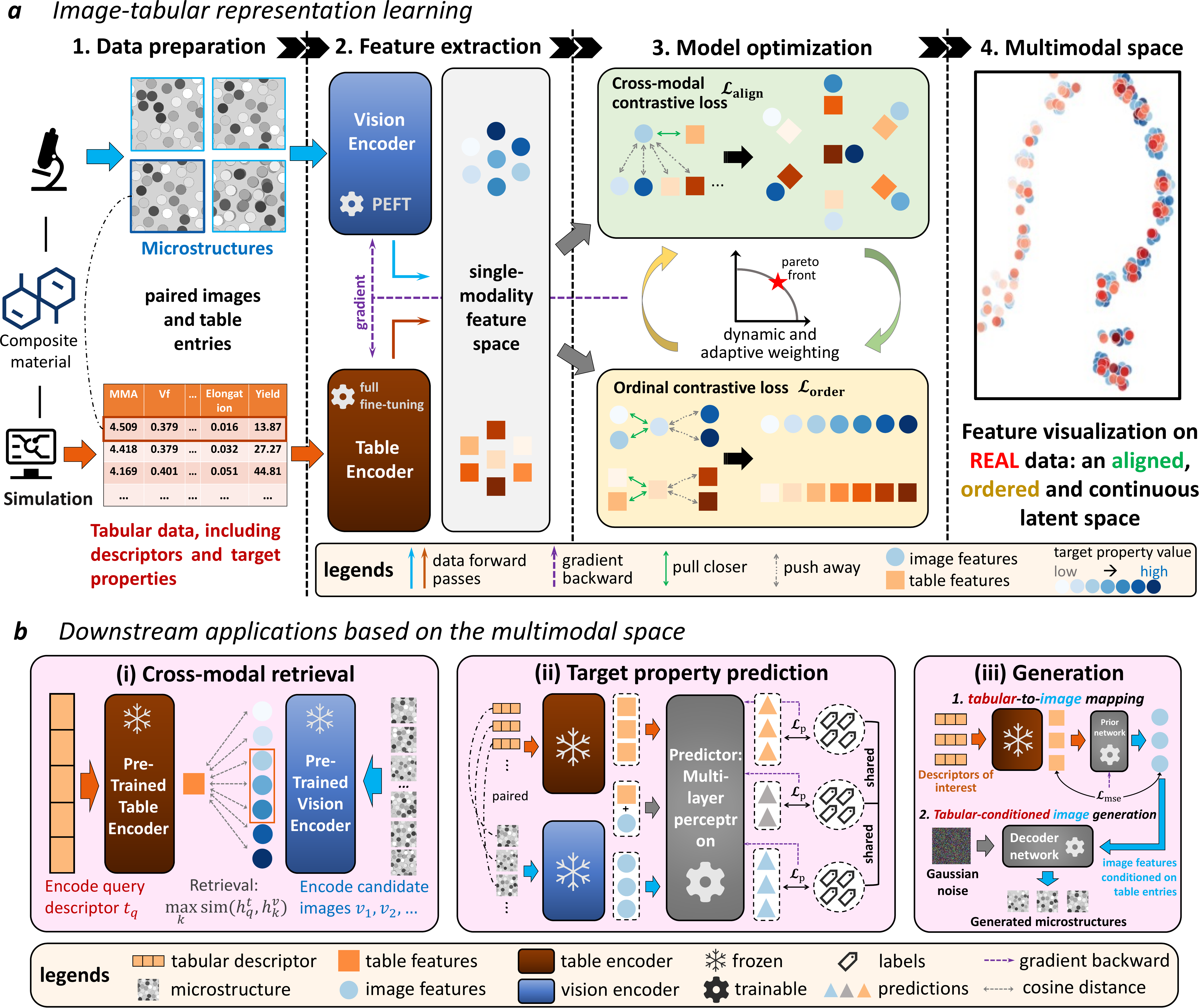}
  \vspace{-6pt}
  \caption{\textbf{a} The pretraining pipeline of ORDER. Step 1, raw composite data images and target properties are obtained by simulating on various descriptors, and organized as pairs. Step 2, paired tabular descriptors and microstructure images are encoded into a shared latent space via dedicated encoders. Step 3, we apply cross-modal contrastive learning \textit{between} modalities to enforce image-tabular alignment, and ordinal-aware contrastive learning \textit{within} each modality to produce property-ordered embeddings. Preference-guided multitask optimization addresses potential conflicts during the bi-objective optimization. Step 4, the training process yields an aligned and ordinal-aware multimodal latent space. \textbf{b} Downstream tasks. The pretrained features are then frozen and serve as the initial starting point for various downstream tasks: (i) Cross-modal retrieval by finding the most similar candidate features with the query feature. The ordinal awareness encoded in features ensures physically meaningful candidates. (ii) Property prediction by training lightweight prediction heads based on the pre-aligned features. (iii) Descriptor-conditioned microstructure generation from tabular inputs. Based on the pretrained feature space, a prior network is trained to translate tabular features into image features and a decoder network learns to reverse them into images.}
  \label{fig_framework}
  \vspace{-8pt}
\end{figure*}   

We present ORDER (ORDinal-aware imagE-tabulaR alignment), a multimodal pretraining framework that not only captures cross-modal relationships but also preserves property ordinality inherent in composite material data. \cref{fig_framework} illustrates the representation learning stage of ORDER and diverse downstream applications based on the pretrained multimodal representations.

The pretraining procedure operates on the Composite datasets described in \cref{sec:dataset}, as well as a public Nanofiber dataset \cite{sgpt}. The inputs include descriptors in tabular form and corresponding microstructural images. Denote the $i_{\rm th}$ input vision microstructure as $v_i$, and its corresponding table feature (descriptor in tabular form) as $t_i$. In the second step of \cref{fig_framework}a, ORDER introduces a vision encoder $E_v$ and table encoder $E_t$ to process the original image and tabular inputs, respectively, and map them to a  shared latent feature space: $h_i^t = E_t(t_i), h_i^v = E_v(v_i)$, where $h_i^t, h_i^v \in \mathbb{R}^{d}$ are $d-$dimension features of tabular and vision modality, respectively.

ORDER pursues two complementary objectives. First, \textit{cross-modal alignment} ensures that feature similarity between matched image-tabular pairs is higher than that of mismatched pairs: $\mathrm{sim}(h_i^t, h_i^v) > \mathrm{sim}(h_i^t, h_j^v)$ for all $i \neq j \in [1,N]$, where $N$ denotes the dataset size. Following established multimodal learning approaches~\cite{radford2021learning}, we employ cross-modal contrastive loss $\mathcal{L}_{\text{align}}$ to achieve this alignment (upper panel of \cref{fig_framework}a, stage 3). This objective pulls closer features from corresponding image-tabular pairs while pushing away mismatched cross-modal pairs and same-modality pairs. However, as illustrated in stage 3 of \cref{fig_framework}a, the standard alignment loss treats all negative pairs equally and ignores the continuous and ordinal characteristics of material property values. Consequently, the resulting feature space may position materials with distinct properties in near regions, potentially confusing downstream tasks.
To address this limitation, ORDER's second objective enforces \textit{ordinal awareness} in the learned representations. Inspired by Zha \textit{et al.}~\cite{rnc}, we apply contrastive loss $\mathcal{L}_{\text{order}}$ within each modality to ensure that samples with similar target properties are embedded closer in the feature space. As depicted in the lower panel of \cref{fig_framework}a (stage 3), this objective produces a continuous and ordered feature distribution aligned with property values. 

\begin{figure*}[!t]
  \centering
  \includegraphics[width=0.85\textwidth]{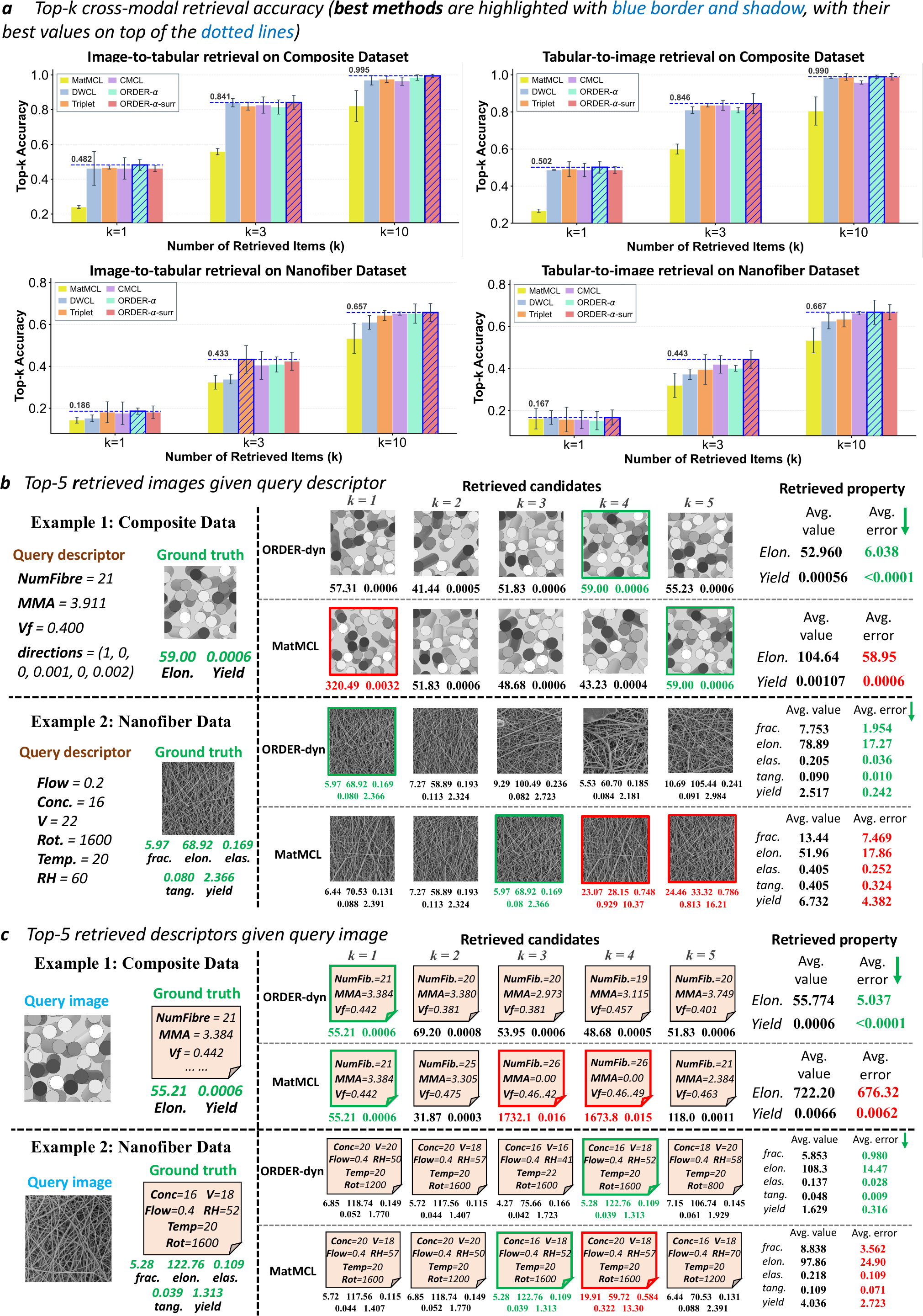}
  \vspace{-6pt}
  \caption{ All results use CLIP-pretrained ViT-B/16. \textbf{a} Top-$k$ cross-modal retrieval accuracy for six methods: MatMCL, DWCL, Triplet, CMCL, ORDER-$\alpha$, and ORDER-$\alpha$-surr. The hatched bar with blue border marks the best-performing method at each $k$. ORDER-$\alpha$ and ORDER-$\alpha$-surr consistently lead, while DWCL and Triplet improve over MatMCL and are competitive with CMCL, confirming the general benefit of continuous-label-aware training. \textbf{b} Examples on top-5 retrieved images given tabular descriptors, shown for ORDER-dyn. The left panel shows query descriptors and the corresponding ground-truth image. The middle panel shows top-5 retrieved examples, with their target property values displayed beneath each retrieved item. Samples with \textcolor{green}{green} borders represent correct retrievals (ground truth), while red borders indicate unwanted candidates with substantially different target properties. The right panel presents statistics of the properties of retrieved samples.  ORDER achieves markedly lower property deviation errors across retrieved candidates. \textbf{c} Examples on top-5 retrieved descriptors given a query image, shown for ORDER-dyn.}
  \label{fig_retrieval}
  \vspace{-6pt}
\end{figure*}

By combining cross-modal alignment and ordinal-aware objectives, ORDER constructs a multimodal latent space that is both semantically aligned and property-ordered. But optimizing these dual objectives can introduce conflicting gradients; to be more specific, cross-modal alignment loss may push apart image features of materials with similar properties, while ordinal-aware loss draws them together. We propose two strategies to mitigate these conflicts: (1) ORDER-$\alpha$ weights the two objectives with a fixed hyperparameter $\alpha$ determined via grid search, and (2) ORDER-dyn dynamically adjust the weights on the losses during training with preference-guided multitask learning~\cite{mahapatra2020multi}. In our experiments, we empirically show that ORDER-$\alpha$ achieves better performance on certain tasks with grid search, while the hyperparameter-free ORDER-dyn exhibits robust performance across diverse downstream  tasks. 
Considering the scarcity of annotations in pretrain data, we further introduce physics-based ordinal surrogates derived from tabular features that preserve inter-sample property ordering. Specifically, the surrogate for the Composite dataset is derived from the Krenchel rule of mixtures using MMA and Vf, while the Nanofiber surrogate is derived from Krenchel orientation efficiency model using mat density and collector-speed-direction interaction. The resultant surrogate-trained variants are termed ORDER-dyn-surr and ORDER-$\alpha$-surr, achieving unsupervised pretraining. The lack of  multimodal materials data has also led previous approaches to initialize vision encoders with unimodal models such as ResNet50~\cite{he2016deep}, constraining model capacity. This work explores the application of pretrained Vision Transformer (ViT)~\cite{dosovitskiy2020image} from CLIP~\cite{radford2021learning}, a large-scale vision-language foundation model. We demonstrate that parameter-efficient fine-tuning (PEFT)~\cite{hu2022lora} enables effective knowledge transfer: the rich pretrained representations in CLIP are preserved while domain-specific materials knowledge is injected using only hundreds of paired samples. For the tabular encoder, we employ FT-Transformer~\cite{gorishniy2021revisiting} following previous work~\cite{sgpt}. We also report results based on ImageNet-pretrained ViT \cite{dosovitskiy2020image} as comparison.
Both encoders are frozen after pretraining and the learned multimodal representations are applied to downstream tasks (\cref{fig_framework}b). Detailed pretraining process can be found in \cref{sec:method_opt}.

We evaluate ORDER's representation quality on our Composite dataset and the public Nanofiber dataset, through diverse downstream tasks including cross-modal retrieval, property prediction, and microstructure generation. The following sections present comprehensive experimental results across these tasks.

\subsection{Cross-modal retrieval results}

Cross-modal retrieval aims to identify the most relevant counterpart in one modality given a query in a different modality, which is fundamental to multimodal materials informatics and inverse design. Without additional fine-tuning, the cosine similarities between ORDER-pretrained features can be computed, as illustrated in \cref{fig_framework}b(i) and detailed in \cref{sec:method_retrieval}. We benchmark ORDER-$\alpha$ and ORDER-$\alpha$-surr against four baselines: (1) vanilla Cross-Modal Contrastive Learning (CMCL), (2) MatMCL~\cite{sgpt}, (3) Distance-Weighted Contrastive Loss (DWCL) \cite{khosla2020supervised}, and (4) property-gap Triplet Loss~\cite{schroff2015facenet}. DWCL and Triplet are customized for this work by adapting standard loss functions to handle real-valued, multidimensional property supervision, which provides a more challenging and informative comparison for ORDER. Details on DWCL and Triplet can be found in \cref{sec:method_baselines}. All methods use identical CLIP-pretrained ViT-B/16 backbones with LoRA. Quantitative Top-$k$ accuracy is shown in \cref{fig_retrieval}a; qualitative retrieval examples using ORDER-dyn are shown in \cref{fig_retrieval}b,c.

\cref{fig_retrieval}a presents quantitative Top-$k$ retrieval accuracy, where retrieval succeeds if the correct counterpart appears among the $k$ candidates. The distance-based baselines outperform MatMCL and CMCL, demonstrating that encoding continuous property information benefits retrieval even without explicit ordinal structure. ORDER-$\alpha$ achieves the best accuracy in most settings, and ORDER-$\alpha$-surr closely matches it despite relying only on physics-derived surrogate labels during pretraining, which validates that surrogates preserve the ordinal structure needed for effective cross-modal alignment. Notably, incorporating intra-modal ordinal-aware contrastive loss may introduce conflicted gradients during cross-modal alignment, potentially limiting raw retrieval accuracy gains. However, this tradeoff is intentional. Our primary objective is not just maximizing Top-$k$ accuracy, but rather constructing a physically meaningful feature space. We therefore assess representation quality through downstream task performance and the physical soundness of retrieved candidates. In practical inverse design scenarios, researchers and practitioners seek to identify multiple candidates with target properties similar to a query sample for subsequent experimental validation. Standard Top-$k$ accuracy metrics only verify ground-truth inclusion but disregard the quality of remaining candidates, which limits its practicability in material applications.

To address the limitation of the current Top-$k$ metric, we analyze the target property distributions of all retrieved items to evaluate their overall quality (\cref{fig_retrieval}b,c). We examine material properties such as yield strength and elongation, which dictate material characteristics and fidelity. Green-bordered samples indicate ground-truth matches, while red-bordered samples deviate substantially from the query. We can observe that, by the definition of Top-$k$ retrieval accuracy ($k=5$ in our case), \textit{all} examples are `correct' retrievals since they include the ground truth. However, the quality of non-ground-truth candidates differs dramatically between ORDER and MatMCL. 
MatMCL frequently retrieves candidates with target properties deviating from the query sample. We show that the average property difference between ground truth and all retrieved items is considerably higher for MatMCL than for ORDER. This discrepancy arises because MatMCL enforces only cross-modal alignment, treating all negative samples equivalently regardless of their ordinal properties. In contrast, ORDER's ordinal-aware objective ensures that materials with similar target properties cluster together in the feature space, yielding retrieval sets where all candidates exhibit relevant property values. The above limitations are also observable in CMCL, suggesting the need of explicit ordinal constraints.

\begin{sidewaystable*}[!tp]
    \caption{Root mean square error (RMSE) of prediction tasks on Composite and Nanofiber datasets and three modalities. Each column refers to one target property to forecast. The results are grouped as modality-specific methods and pretraining methods with different vision backbones. Best results within each group are bolded. The \textit{Avg. Rank} column reports the average competition rank of each pretraining method across all 21 evaluation cases within its backbone group. Lower rank is better.}
    \label{tab_predict}
    \vspace{-4pt}
    \centering
    \resizebox{0.99\linewidth}{!}{
    \setlength{\tabcolsep}{4pt}
    \begin{tabular}{lccccccc|lccccccc|lccccccc|c}
    \toprule
    \multicolumn{8}{c}{Tabular features} & \multicolumn{8}{|c}{Image features} & \multicolumn{8}{|c}{Fusion features} & \multicolumn{1}{|c}{\multirow{3}{*}{\shortstack{Avg.\\Rank}}} \\ \cmidrule{1-24}
    \multirow{2}{*}{Method} & \multicolumn{5}{c}{Nanofiber} & \multicolumn{2}{c|}{Composite} &
    \multirow{2}{*}{Method} & \multicolumn{5}{c}{Nanofiber} & \multicolumn{2}{c|}{Composite} &
    \multirow{2}{*}{Method} & \multicolumn{5}{c}{Nanofiber} & \multicolumn{2}{c}{Composite} & \multicolumn{1}{|c}{} \\
    & frac. & elon. & elastic & tangent & yield & Yield & Elon. &
     & frac. & elon. & elastic & tangent & yield & Yield & Elon. &
     & frac. & elon. & elastic & tangent & yield & Yield & \multicolumn{1}{c|}{Elon.} \\ \hline
    \rowcolor[HTML]{D9D9D9}
    \multicolumn{25}{l}{\textit{Modality-specific baselines}} \\
    TabPFN~\cite{hollmann2025accurate} & 2.723 & 21.495 & \textbf{0.115} & \textbf{0.099} & \textbf{1.697} & \textbf{234.589} & \textbf{0.0021} &
    ResNet50~\cite{he2016deep} & 2.441 & 22.927 & 0.133 & 0.106 & 1.871 & 360.686 & 0.0030 &
    \multicolumn{8}{c|}{---} & --- \\
    XGBoost~\cite{chen2016xgboost} & \textbf{2.638} & 21.536 & 0.133 & 0.100 & 1.761 & 278.410 & 0.0030 &
    ViT-B/16~\cite{dosovitskiy2020image} & 2.410 & \textbf{21.799} & \textbf{0.124} & \textbf{0.096} & \textbf{1.722} & \textbf{247.635} & \textbf{0.0020} &
    \multicolumn{8}{c|}{} & --- \\
    CatBoost~\cite{prokhorenkova2018catboost} & 2.666 & \textbf{20.199} & 0.122 & 0.113 & 1.745 & 283.365 & 0.0025 &
    ResNet101~\cite{he2016deep} & \textbf{2.381} & 22.037 & 0.127 & 0.097 & 1.811 & 425.413 & 0.0032 &
    \multicolumn{8}{c|}{} & --- \\
    LightGBM~\cite{ke2017lightgbm} & 3.231 & 23.674 & 0.146 & 0.131 & 2.293 & 264.710 & 0.0023 &
    ViT-B/32~\cite{dosovitskiy2020image} & 2.457 & 22.872 & 0.125 & 0.108 & 1.806 & 341.440 & 0.0026 &
    \multicolumn{8}{c|}{} & --- \\
    \rowcolor[HTML]{D9D9D9}
    \multicolumn{25}{l}{\textit{Multimodal pretraining --- CLIP-pretrained ViT-B/16}} \\
    MatMCL~\cite{sgpt} & 2.772 & 21.704 & 0.118 & 0.106 & 1.992 & 277.157 & 0.0022 &
    MatMCL & 2.536 & 22.900 & 0.132 & 0.109 & 1.906 & 246.621 & 0.0020 &
    MatMCL & 2.433 & 20.050 & 0.118 & 0.097 & 1.768 & 290.335 & 0.0024 & 6.81 \\
    CMCL~\cite{radford2021learning} & 2.966 & 21.649 & 0.125 & 0.112 & 1.944 & 279.757 & 0.0022 &
    CMCL & 2.505 & 23.359 & 0.137 & 0.107 & 1.839 & 298.337 & 0.0024 &
    CMCL & 2.556 & 21.218 & 0.123 & 0.100 & 1.809 & 297.189 & 0.0023 & 7.33 \\
    DWCL\cite{khosla2020supervised} & 2.341 & 20.624 & 0.100 & 0.095 & 1.624 & 245.119 & 0.0020 &
    DWCL & 2.321 & 22.892 & 0.125 & 0.103 & 1.727 & 226.828 & 0.0019 &
    DWCL & 2.026 & 19.827 & \textbf{0.089} & 0.086 & \textbf{1.304} & 225.427 & \textbf{0.0018} & 2.90 \\
    Triplet~\cite{schroff2015facenet} & 2.447 & 21.946 & 0.105 & 0.101 & 1.874 & 249.860 & 0.0019 &
    Triplet & 2.235 & 24.040 & 0.126 & 0.102 & 1.729 & 229.312 & 0.0019 &
    Triplet & \textbf{1.892} & 20.455 & 0.091 & 0.082 & 1.498 & 232.288 & \textbf{0.0018} & 4.10 \\
    \rowcolor[RGB]{221,235,247}
    ORDER-dyn & 2.537 & 20.792 & 0.109 & 0.106 & 1.763 & 221.346 & \textbf{0.0018} &
    ORDER-dyn & 2.289 & 23.418 & 0.127 & 0.102 & 1.721 & 226.944 & 0.0019 &
    ORDER-dyn & 2.152 & 19.436 & 0.097 & 0.088 & 1.519 & 223.815 & 0.0019 & 3.76 \\
    \rowcolor[RGB]{221,235,247}
    ORDER-$\alpha$ & \textbf{2.335} & 21.201 & 0.102 & \textbf{0.092} & 1.632 & 234.170 & \textbf{0.0018} &
    ORDER-$\alpha$ & \textbf{2.219} & 24.369 & \textbf{0.120} & 0.103 & 1.703 & 227.028 & \textbf{0.0018} &
    ORDER-$\alpha$ & 1.980 & \textbf{19.153} & 0.093 & \textbf{0.079} & 1.512 & 226.431 & \textbf{0.0018} & \textbf{2.76} \\
    \rowcolor[RGB]{221,235,247}
    ORDER-dyn-surr & 2.503 & \textbf{19.480} & \textbf{0.093} & 0.099 & \textbf{1.545} & \textbf{213.571} & 0.0019 &
    ORDER-dyn-surr & 2.320 & 23.574 & 0.130 & \textbf{0.101} & \textbf{1.690} & \textbf{213.593} & 0.0019 &
    ORDER-dyn-surr & 2.264 & 19.472 & 0.102 & 0.089 & 1.438 & \textbf{212.002} & 0.0019 & 3.05 \\
    \rowcolor[RGB]{221,235,247}
    ORDER-$\alpha$-surr & 2.442 & 21.729 & 0.098 & 0.102 & 1.593 & 220.131 & \textbf{0.0018} &
    ORDER-$\alpha$-surr & 2.390 & \textbf{22.782} & 0.127 & 0.106 & 1.745 & 227.202 & 0.0020 &
    ORDER-$\alpha$-surr & 2.169 & 21.225 & 0.106 & 0.092 & 1.409 & 218.884 & 0.0019 & 4.24 \\ \hline
    \rowcolor[HTML]{D9D9D9}
    \multicolumn{25}{l}{\textit{Multimodal pretraining --- ImageNet-pretrained ViT-B/16}} \\
    MatMCL~\cite{sgpt} & 2.766 & 21.401 & 0.124 & 0.112 & 1.878 & 260.404 & \textbf{0.0020} &
    MatMCL & 2.546 & 23.547 & 0.132 & 0.107 & 1.855 & 276.692 & 0.0023 &
    MatMCL & 2.475 & 20.683 & 0.128 & 0.105 & 1.843 & 294.815 & 0.0024 & 6.71 \\
    CMCL~\cite{radford2021learning} & 2.762 & 21.608 & 0.121 & 0.109 & 1.770 & 320.823 & 0.0026 &
    CMCL & 2.369 & 22.479 & 0.132 & \textbf{0.099} & 1.801 & 300.240 & 0.0025 &
    CMCL & 2.245 & 21.200 & 0.102 & 0.092 & 1.465 & 291.008 & 0.0023 & 6.10 \\
    DWCL~\cite{khosla2020supervised} & 2.390 & 21.357 & 0.103 & \textbf{0.098} & 1.690 & 274.662 & 0.0022 &
    DWCL & 2.361 & 24.040 & \textbf{0.124} & 0.107 & 1.752 & 230.856 & \textbf{0.0019} &
    DWCL & 2.151 & 19.783 & 0.093 & 0.087 & 1.414 & 241.714 & 0.0019 & 3.76 \\
    Triplet~\cite{schroff2015facenet} & \textbf{2.389} & 22.054 & 0.103 & 0.099 & 1.758 & 264.699 & 0.0021 &
    Triplet & 2.307 & 23.672 & 0.135 & \textbf{0.099} & 1.738 & 217.789 & \textbf{0.0019} &
    Triplet & \textbf{2.001} & 22.462 & \textbf{0.092} & 0.084 & \textbf{1.393} & 213.877 & \textbf{0.0018} & 3.24 \\
    \rowcolor[RGB]{221,235,247}
    ORDER-dyn & 2.623 & \textbf{19.292} & 0.105 & 0.102 & 1.757 & 266.089 & 0.0021 &
    ORDER-dyn & 2.339 & 23.837 & 0.131 & 0.106 & 1.777 & 224.526 & \textbf{0.0019} &
    ORDER-dyn & 2.334 & 20.306 & 0.097 & 0.091 & 1.580 & 225.957 & \textbf{0.0018} & 4.29 \\
    \rowcolor[RGB]{221,235,247}
    ORDER-$\alpha$ & 2.451 & 20.706 & 0.111 & \textbf{0.098} & 1.732 & 267.971 & \textbf{0.0020} &
    ORDER-$\alpha$ & \textbf{2.267} & 23.790 & 0.127 & 0.102 & \textbf{1.731} & 225.587 & \textbf{0.0019} &
    ORDER-$\alpha$ & 2.244 & \textbf{19.659} & 0.097 & \textbf{0.082} & 1.572 & 232.135 & \textbf{0.0018} & \textbf{3.05} \\
    \rowcolor[RGB]{221,235,247}
    ORDER-dyn-surr & 2.484 & 19.399 & \textbf{0.099} & 0.099 & \textbf{1.505} & 218.777 & \textbf{0.0020} &
    ORDER-dyn-surr & 2.391 & \textbf{22.430} & 0.133 & \textbf{0.099} & 1.826 & 217.204 & 0.0020 &
    ORDER-dyn-surr & 2.353 & 20.483 & 0.102 & 0.088 & 1.577 & 214.398 & 0.0020 & 3.71 \\
    \rowcolor[RGB]{221,235,247}
    ORDER-$\alpha$-surr & 2.558 & 19.909 & 0.100 & 0.102 & 1.590 & \textbf{216.195} & \textbf{0.0020} &
    ORDER-$\alpha$-surr & 2.364 & 23.082 & 0.130 & 0.102 & 1.789 & \textbf{216.104} & 0.0020 &
    ORDER-$\alpha$-surr & 2.268 & 20.706 & 0.102 & 0.091 & 1.437 & \textbf{213.545} & 0.0020 & 3.57 \\ \hline
    \end{tabular}}
    \vspace{-6pt}
\end{sidewaystable*}

\subsection{Property prediction performances}
\label{sec:prediction}
\begin{figure*}[!t]
  \centering
  \includegraphics[width=0.86\textwidth]{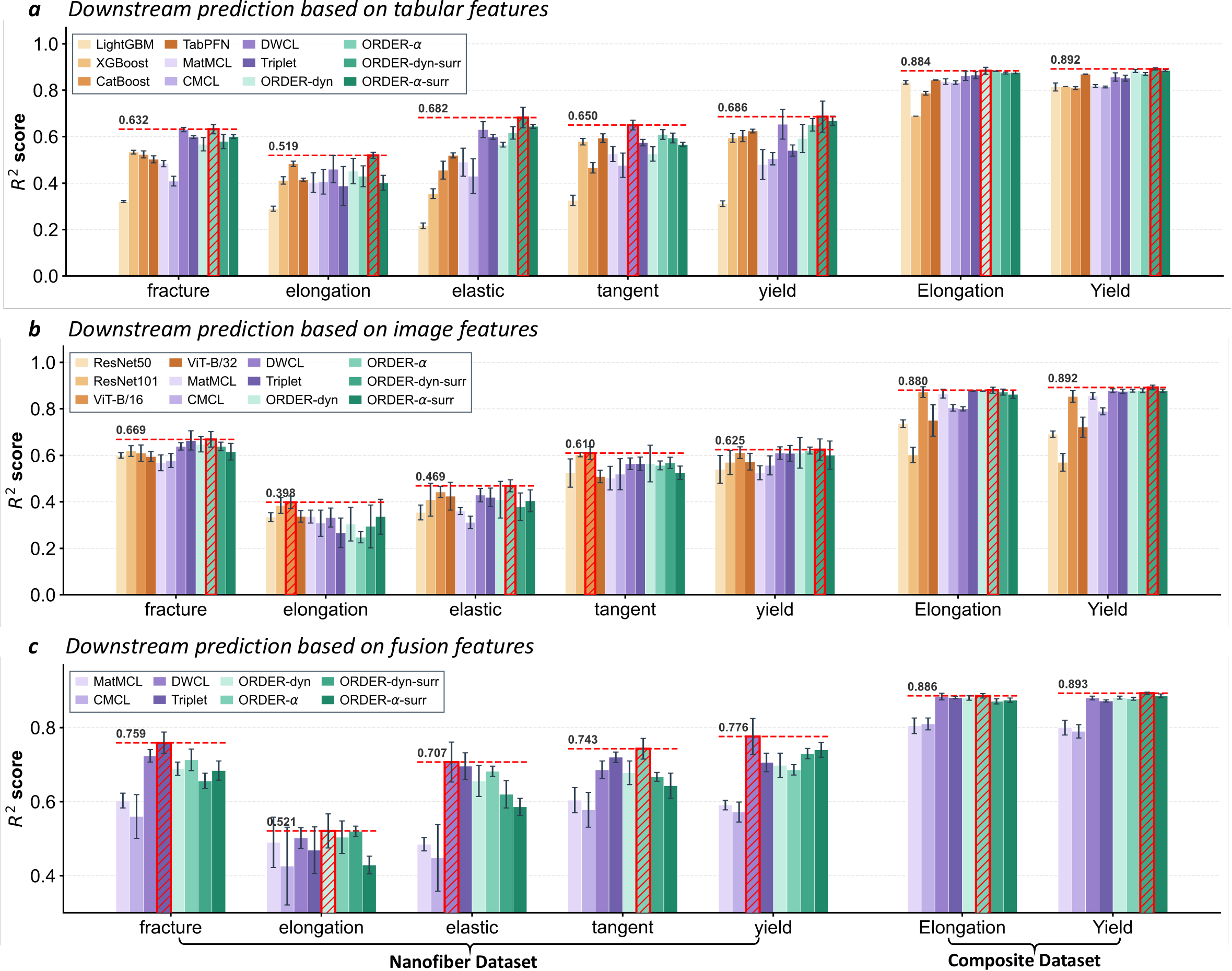}
  \vspace{-9pt}
  \caption{$R^2$ scores for property prediction on Nanofiber and Composite datasets, shown separately for tabular, image, and modality fused features. Each bar group stands for one target property. Within each group, bars are sorted by method type: modality-specific baselines in yellow, multimodal pretraining baselines in purple (MatMCL, CMCL, DWCL, Triplet), and ORDER variants in green. The hatched bar with red border marks the best-performing method per property. All experiments use CLIP-pretrained ViT-B/16 as backbone.}
  \label{fig_predict}
  \vspace{-12pt}
\end{figure*}

Accurately predicting target material characteristics from input features is one of the essential tasks in materials science, as it enables materials design by bypassing costly experimental characterization. This task typically operates on tabular data, where established tabular-specific methods such as XGBoost~\cite{chen2016xgboost}, TabPFN~\cite{hollmann2025accurate}, CatBoost~\cite{prokhorenkova2018catboost} and LightGBM~\cite{ke2017lightgbm} have demonstrated strong performance. As ORDER is a multimodal framework involving image and tabular modalities, we additionally evaluate vision-based prediction capabilities by comparing against general vision models including ImageNet-pretrained ResNet50, ResNet101~\cite{he2016deep} and Vision Transformer (ViT-B/16, ViT-B/32)~\cite{dosovitskiy2020image}. We also benchmark against the two distance-based contrastive baselines DWCL~\cite{khosla2020supervised} and Triplet~\cite{schroff2015facenet}, as well as surrogate-trained ORDER variants ORDER-dyn-surr and ORDER-$\alpha$-surr, which substitute physics-based ordinal proxies for ground-truth property labels during pretraining. For modality-specific baselines, we train or fine-tune models directly on raw inputs ($t_i$, $v_i$). For multimodal pretraining methods (ORDER, MatMCL, CMCL, DWCL, Triplet), encoders remain frozen after pretraining, and pre-extracted features ($h_i^t$, $h_i^v$) are used to train a lightweight multilayer perceptron (MLP) for property prediction.  We evaluate all methods on predicting fracture strength, elongation, elastic modulus, tangent modulus, and yield strength for the Nanofiber dataset, and yield strength and elongation for the Composite dataset. All experiments are repeated five times independently with mean values reported. Results are presented in \cref{tab_predict} (RMSE) and \cref{fig_predict} (R$^2$ scores).

For tabular-based prediction on the Nanofiber dataset, ORDER-dyn-surr achieves the lowest RMSE for elongation, elastic modulus, and yield strength among all pretraining-based methods, while ORDER-$\alpha$ attains the best fracture strength and tangent modulus error. Specialized tabular models (TabPFN, CatBoost) remain strong on individual properties but are outperformed by ORDER variants across the full property set. On the Composite dataset, ORDER-dyn-surr achieves the lowest yield-strength RMSE in the tabular modality (213.6), substantially below CMCL (279.8) and the best modality-specific baseline TabPFN (234.6).
For image-based prediction, our proposed ORDER-dyn-surr and ORDER-$\alpha$ achieve the lowest or near-lowest composite yield-strength RMSE across all three modalities, confirming that ordinal-aware pretraining produces better image representations than general vision models and pretraining baselines.

\cref{tab_predict} also compares ORDER variants against other distance-based baselines DWCL~\cite{khosla2020supervised} and Triplet~\cite{schroff2015facenet} that encode continuous property structure directly. We can observe that ORDER variants demonstrate clear superiority on both datasets and backbones across tabular and image-feature-based prediction, achieving the lowest error in 25/28 cases. On the Nanofiber dataset, DWCL and Triplet achieve competitive performance on certain properties (e.g., Triplet attains the best fusion-modality fracture strength accuracy), highlighting the general value of distance-aware training. Nonetheless, ORDER maintains an overall advantage through its ordinal structure across multi-property scenarios. The \textit{Avg. Rank} column in \cref{tab_predict} further supports this conclusion: ORDER-$\alpha$ achieves the best average rank in both backbone configurations, confirming that its superiority is broad and consistent rather than concentrated on a few properties. Among the compared pretraining baselines, MatMCL and CMCL rank last in both groups, whereas distance-based methods DWCL and Triplet are more competitive but still behind ORDER-$\alpha$ in both settings. Such results underscore the benefit of ordinal structure over soft-weighting or fixed-margin training objectives.

The surrogate-trained variants ORDER-dyn-surr and ORDER-$\alpha$-surr achieve performance matching or exceeding their ground-truth-label counterparts in many settings. Most notably, ORDER-dyn-surr achieves the best CLIP-backbone RMSE for Composite yield strength under all three modalities, and leads on several Nanofiber properties including elongation and yield strength. This demonstrates that the physics-based surrogates preserve the ordinal structure needed for effective pretraining, validating a practical path to label-efficient materials representation learning.

We observe that prediction performances based on different modalities differ across target properties and datasets, consistent with our claim that both descriptors and microstructure images are essential. On our Composite dataset with coarse-level descriptors (only fiber density and misalignment angle), image-based predictions capture additional spatial information invisible to the tabular branch. The richer descriptors in the Nanofiber dataset make tabular features competitive or superior for some properties (e.g., elongation). Therefore, we also report multimodal fusion (fusion technique detailed in \cref{sec:method_predict}) results in \cref{tab_predict}. We can observe that modality fusion enhances distance-based baselines to achieve competitive performance with our ORDER series, and all fusion results consistently outperforms single-modality predictions results.  \cref{tab_predict} also reports results for both CLIP-pretrained and ImageNet-pretrained ViT-B/16 backbones to understand the contributions made by different pretraining objectives. CLIP-backbone results generally outperform their ImageNet counterparts, reflecting the richer initial representations in the CLIP vision transformer, while both backbones confirm the consistent advantage of ORDER over the compared image-tabular alignment baselines.

\begin{figure*}[!t]
  \centering
  \includegraphics[width=0.9\textwidth]{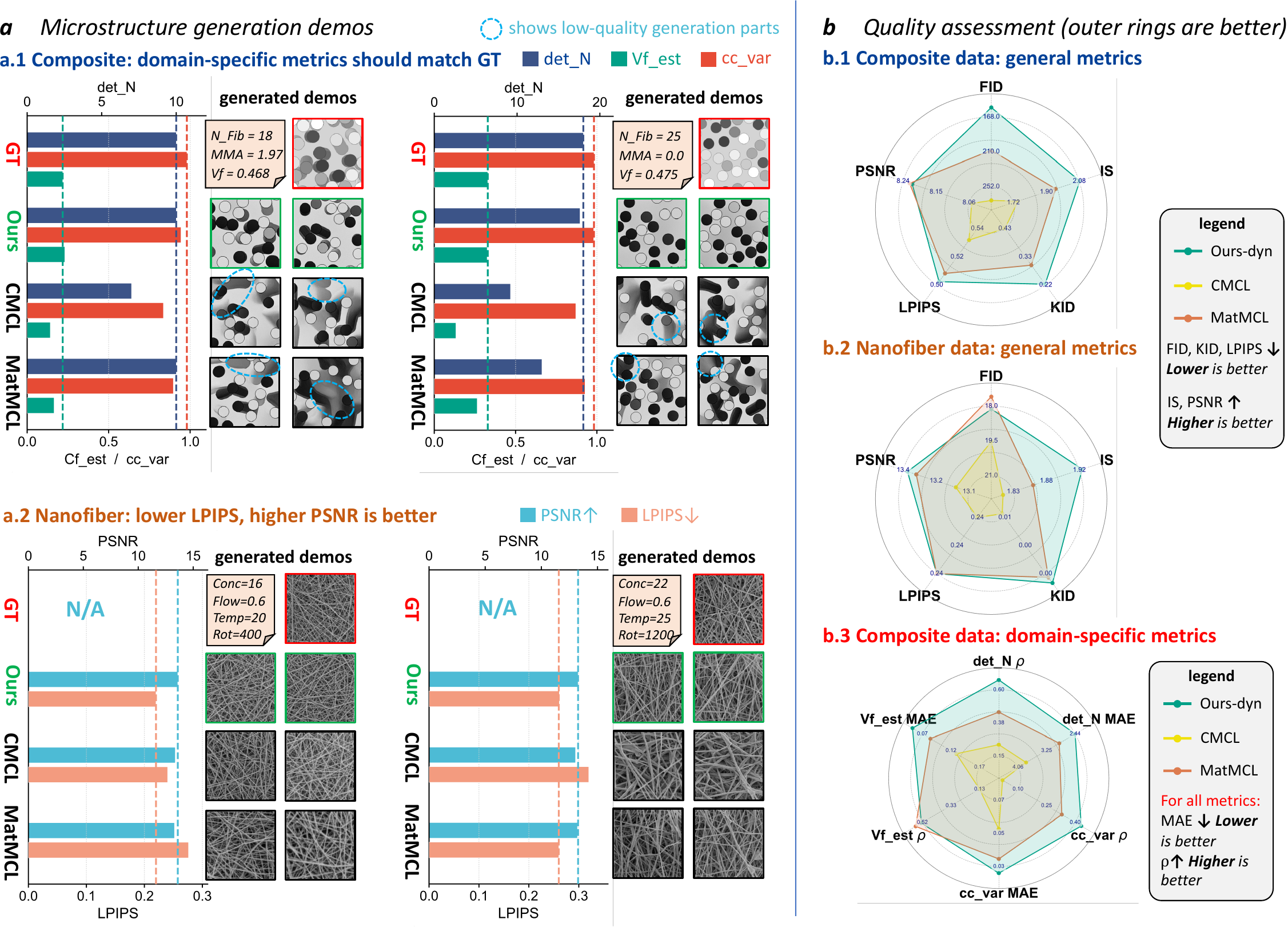}
  \vspace{-10pt}
  \caption{Descriptor-conditioned microstructure generation. \textbf{a} Representative generation examples from ORDER-dyn (Ours), CMCL, and MatMCL. Two randomly selected generation demos are shown per method. In adjcent bar charts the four horizontal groups correspond to ground truth (GT) and the three compared methods, where in \textbf{a.1 Composite data} three bars per group are domain-specific evaluation metrics that are expected to align with the GT ones (indicated by dotted lines). \textbf{a.2 Nanofiber data} include two bars per group showing PSNR and LPIPS for the generated demos, and dashed lines indicate the best result per metric across methods. \textbf{b} Radar-chart quantitative evaluation, where outer rings indicate better performance. \textbf{b.1-b.2} includes standard image-quality metrics (FID, KID, LPIPS, IS, and PSNR) for Composite and Nanofiber data. \textbf{b.3} includes our proposed domain-specific evaluation for Composite data. ORDER-dyn achieves consistent improvements over multimodal baselines across both standard and domain-specific evaluations.}
  \label{fig_generate}
  \vspace{-10pt}
\end{figure*}

\cref{fig_predict} presents $R^2$ scores across three panels (tabular-, image-, fusion-based prediction). Comparing the three panels, fusing both modalities consistently raises $R^2$ for every method and every property, confirming that tabular descriptors and microstructure images carry complementary information. Across all three modalities, ORDER variants rank among the top methods on the majority of properties (being the best method in 15/21 cases). Notably, ORDER-dyn-surr achieves the highest Composite Yield $R^2$ in all three modality settings, while ORDER-$\alpha$ leads on Composite Elongation in the fusion panel. For the Nanofiber properties, DWCL and Triplet are competitive on a subset of properties, reflecting the general benefit of distance-aware training.  In the image panel, vanilla vision models (ViT-B/16) remain competitive for elongation, where the visual texture carries particularly strong signals. The surrogate-trained ORDER-dyn-surr and ORDER-$\alpha$-surr match or exceed their standard counterparts across most settings, especially for Composite Yield. We deduce the reason is that surrogate signals reduce potential overfitting caused by ground-truth ordering.

The performance gains achieved by ORDER stem from its ordinal-aware feature distributions. Intuitively, prediction becomes considerably easier when input features are pre-organized according to target properties, enabling the MLP to exploit continuous property-feature relationships and interpolate between sparsely unobserved samples. This ordinal structure is injected during pretraining via ordinal-aware contrastive learning without compromising cross-modal alignment, as evidenced by ORDER's competitive retrieval accuracy in \cref{fig_retrieval}. The surrogate experiments further reveal that such ordinal organization does not strictly require ground-truth labels: physics-grounded proxies that preserve inter-sample ranking are sufficient,  extending the practical reach of ORDER. 

\begin{figure*}[!t]
  \centering
  \includegraphics[width=0.95\textwidth]{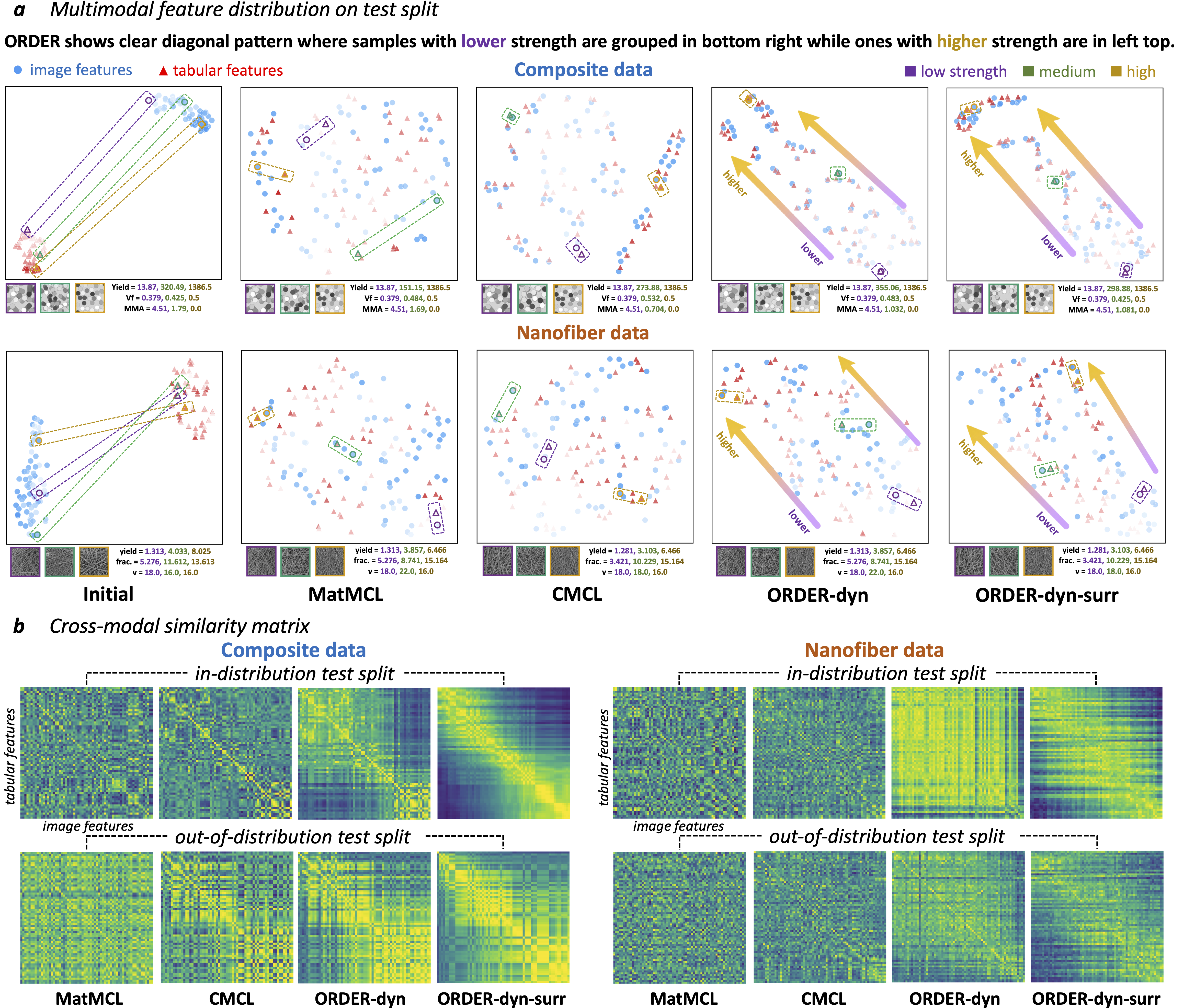}
  \vspace{-8pt}
    \caption{Analysis of multimodal representations using yield strength as target property. \textbf{a} T-SNE projections of the learned feature space for Composite (top) and Nanofiber (bottom) data. Blue circles denote image features and red triangles denote tabular features. Light to dark point color indicates for low to high strength. Below each t-SNE plot, three representative microstructure thumbnails are shown for samples with low, mid, and high property values alongside their descriptor details. The image-tabular pairs of the thumbnail samples are highlighted in the t-SNE figures. \textbf{b} Cross-modal feature similarity matrices for samples sorted by target property values, for Composite (left) and Nanofiber (right) data. Each panel contains two rows: in-distribution test split (top) and out-of-distribution (OOD) test split (bottom), where OOD splits include samples with property ranges not seen during training. Brighter color indicates higher cross-modal similarity.}
  \label{fig_analysis}
  \vspace{-8pt}
\end{figure*}

\subsection{Descriptor-conditioned microstructure generation}
\label{sec:generation}
Image-based generation visualizes the material design process, revealing fiber orientation effects and local microstructures to facilitate defect identification and design optimization. Compared with single tabular modality, image analysis can directly capture more specific spatial fiber distribution. Acquiring high-quality microstructure images generally requires laboring and expensive experiment or simulations. Leveraging ORDER's property-aware multimodal latent space, we can generate realistic microstructures conditioned on input tabular descriptors. Our generation experiments adopt the hyperparameter-free ORDER-dyn to meet practical requirements, since optimal $\alpha$ is hard to obtain in advance when generating unseen structures.

We adopt a two-stage generative framework following DALLE-2~\cite{ramesh2022hierarchical}. As illustrated in \cref{fig_framework}b(iii), the pipeline includes a prior network and a decoder trained on pre-extracted ORDER features. The prior network maps tabular descriptors to their corresponding image feature distributions, while the decoder synthesizes microstructures from image features using a diffusion-based generative process (detailed in \cref{sec:method_generation}). The generative architecture and training procedure remain identical across all compared methods, with only the quality of pretrained multimodal features differs. Therefore, the generation results can directly reflect the representation learning effectiveness.

We evaluate generated microstructure quality using two complementary approaches. Standard image-quality metrics: FID~\cite{fid}, KID~\cite{kid}, LPIPS~\cite{lpips}, IS~\cite{is}, and PSNR~\cite{psnr} capture distributional similarity, perceptual quality, and pixel-level accuracy. For composite materials microstructures, however, these metrics are insufficient: a generated image may achieve competitive FID or LPIPS while producing incorrect fiber counts or misalignment patterns. These defects are invisible to generic metrics but critical for property prediction and structural design. We therefore design three domain-specific structural metrics that directly quantify physical fidelity of composite microstructures: detected fiber count (det\_N), gradient circular variance (cc\_var), and contour area fraction (Vf\_est), corresponding to NumFibers, MMA, and Vf, respectively. Each metric is validated against ground-truth tabular labels on real composite images (more details in \cref{sec:method_generation}).

\cref{fig_generate}a presents representative generated examples comparing ORDER-dyn with baseline methods CMCL and MatMCL. On the Nanofiber dataset, ORDER-dyn generates microstructures that reproduce the visual characteristics and fiber density distributions of ground-truth images, while CMCL and MatMCL exhibit degraded resolution, inconsistent fiber density, and reduced sharpness. On the Composite dataset, ORDER-dyn generates fiber arrangements that closely match input descriptors in fiber count and misalignment angle, whereas CMCL and MatMCL produce structures with irregular shapes, spurious shadows, and inconsistent morphologies. As highlighted by blue dotted circles in the figure, CMCL and MatMCL might generate incomplete or blurred fibers (left panel) or wrongly show fiber's cylinder body when MMA is 0 (right panel). These observations are directly quantified by the adjacent bar charts. For Composite data (\cref{fig_generate}a.1) the bars represent our proposed domain-specific metrics, where dashed reference lines mark the GT values to match. ORDER-dyn's bars align closely to GT values, confirming structural accuracy across all three descriptors, while CMCL and MatMCL show systematic deviations. For Nanofiber data (\cref{fig_generate}a.2) the bars show sample-wise generation quality metric PSNR (higher is better) and LPIPS (lower is better), where ORDER-dyn consistently achieves the best performances.

Radar charts in \cref{fig_generate}b.1-b.2 present quantitative evaluation using all five standard metrics across both datasets, where outer rings indicate better performance. ORDER-dyn achieves consistently strong results. On Composite data (\cref{fig_generate}b.1), MatMCL attains comparable PSNR scores but underperforms on distributional similarity (FID, KID), perceptual quality (LPIPS), and diversity (IS), indicating that pixel-level accuracy alone does not guarantee structural fidelity. On Nanofiber data (\cref{fig_generate}b.2), ORDER-dyn achieves similar and slightly better performance then MatMCL on FID, PSNR, LPIPS, KID, and a significant improvement on IS. CMCL shows greater degradation on all metrics. \cref{fig_generate}b.3 further reports domain-specific metrics on the Composite dataset. Here Spearman correlation $\rho$ measures whether computed metrics in generated images matches that in real images, and mean absolute error (MAE) measures the per-sample absolute deviation of the detected metrics from the ground-truth ones. ORDER-dyn achieves the highest $\rho$ for det\_N and cc\_var, and the lowest MAE across all three structural metrics. MatMCL slightly surpasses ORDER-dyn on Vf\_est $\rho$, suggesting it preserves relative volume-fraction ordering across conditions. However, its substantially higher MAE reveals a systematic offset in absolute packing density, and it falls below ORDER-dyn on all other metrics. The advantage of ORDER is pronounced for cc\_var, where CMCL shows near-zero correlation, indicating failure to reproduce the misalignment signature. Together, these results confirm that ORDER's ordinal-aware latent space produces microstructures with more accurate physical structure, not merely improved pixel statistics.

These quantitative results support the qualitative observations in \cref{fig_generate}a. By ensuring that latent features vary continuously with target properties, ORDER produces physically meaningful and effective priors that interpolate between observed material configurations. Such priors contribute to targeted generation of candidate microstructures with desired physical characteristics even for unseen design conditions.

\subsection{Method analysis}
\label{sec:analysis}
To validate the effects of dual optimization objectives, we visualize the multimodal feature spaces learned by ORDER and baseline methods on the test split. \cref{fig_analysis}a presents t-SNE~\cite{maaten2008visualizing} projections comparing five conditions: Initial (before pretraining), MatMCL, CMCL, ORDER-dyn, and ORDER-dyn-surr, for both Composite and Nanofiber datasets. Blue circles denote image features and red triangles denote tabular features, with point color encoding property level from light (low) to dark (high).  Before pretraining (`Initial'), image and tabular features occupy separate regions of the embedding space. After pretraining, ORDER-dyn and ORDER-dyn-surr achieve superior cross-modal alignment on both datasets, with paired blue and red points overlapping. In contrast, MatMCL and CMCL exhibit incomplete alignment with partially separated modality clusters.
Ordinal-aware contrastive learning further organizes the feature space by target property values, as indicated by the directional arrows in \cref{fig_analysis}a. On the simpler Composite dataset, ORDER-dyn and ORDER-dyn-surr achieve a clear and consistent low-to-high property gradient across the embedding space, whereas MatMCL and CMCL show weaker or inconsistent property-based organization. On the more challenging Nanofiber dataset, ORDER-dyn and ORDER-dyn-surr again form coherent property-driven structures with directional gradient, while MatMCL and CMCL show no systematic property-based organization. The representative microstructure thumbnails at low, mid, and high property levels confirm that these ordinal structures correspond to physically meaningful differences in fiber spatial arrangement. These t-SNE visualizations demonstrate that ORDER successfully constructs an aligned and property-ordered multimodal space.

\cref{fig_analysis}b presents cross-modal feature similarity matrices for samples \textbf{sorted by target property values}, evaluated on both in-distribution and out-of-distribution (OOD) test splits, where OOD splits comprise samples with property ranges not seen during training. MatMCL and CMCL produce  sharp diagonal patterns, indicating high similarity only for directly matched image-tabular pairs while largely ignoring property relationships among nearby samples. In contrast, ORDER-dyn and ORDER-dyn-surr exhibit not only strong diagonal signals but also broad off-diagonal bands of elevated similarity that decay smoothly with increasing property distance, reflecting their ordinal-aware embedding structure. ORDER-dyn-surr maintains similarity patterns comparable to ORDER-dyn and show smoother changing patterns. On OOD test splits, these similarity patterns weaken for all methods due to distribution shift, but ORDER-dyn and ORDER-dyn-surr continue to exhibit clearer off-diagonal structure than MatMCL and CMCL on both datasets. These cross-modal similarity patterns collaborate with the t-SNE analysis to confirm that ORDER effectively constructs aligned and property-ordered multimodal representations. This visualization analysis also provides insights into ORDER's superior downstream performance across retrieval (\cref{fig_retrieval}), prediction (\cref{fig_predict}), and generation (\cref{fig_generate}) tasks.

\section{Discussion}
\label{sec:discussion}
This work focuses on composite materials, whose material structure and properties are decided by fiber distributions in a continuous design space. Therefore, the discrete graph structures widely used to describe crystals and polymers fail to model composite materials. These challenges motivate our design of a vision-tabular pretraining framework for composite material to  encode both tabular descriptors and their corresponding microstructural images. Current multimodal pretraining approaches deteriorate for composites with continuous, infinite-dimensional design spaces and extreme data scarcity. To address this, we propose ORDinal-aware imagE-tabulaR alignment (ORDER) that constructs aligned and property-ordered multimodal representations. ORDER preserves the continuous nature of composite design spaces and interpolates across unobserved design spaces with merely hundreds of data points.

The core value of ORDER is the learned materials-physics-aware representations. The tabular encoder $E_t$ processes design-level physical parameters and maps them to a feature vector whose geometric position reflects the material's position in the continuous design space. The image encoder $E_v$ extracts spatial microstructural cues corresponding to mechanical properties of the material and maps them to the shared latent space with tabular features. The potential of these representations is evaluated on a public Nanofiber-reinforced composite dataset and our in-house CF-T700 Composite dataset. In cross-modal retrieval, ORDER achieves not only competitive accuracy but retrieves candidates with substantially lower property deviation, directly enhancing practical utility for inverse design where multiple viable candidates are needed. For property prediction, ORDER variants achieve consistently lower error than modality-specific models and multimodal baselines across both datasets. ORDER variants further outperform distance-based contrastive baselines (DWCL and Triplet) that also incorporate continuous property information. This indicates that ORDER's rank-based ordinal structure provides a more effective inductive bias than soft-weighting or fixed-margin approaches.  Averaged across all 21 evaluation cases, ORDER-$\alpha$ achieves the best overall rank in both the CLIP-pretrained and ImageNet-pretrained groups, confirming its consistent and robust gains. Facing the annotation scarcity of target signals during pretraining, we introduce physics-based ordinal surrogates that replace ground-truth property labels with feature-derived proxies. The surrogate framework generalizes to any domain where a constitutive or empirical model provides a physics-grounded ranking of samples. The surrogate-trained ORDER variants  achieve performance matching or exceeding their ground-truth-label counterparts in many settings, validating that the ordinal structure can be recovered from domain knowledge without complete property supervision. 
Microstructure generation on unseen data further confirms the physical significance of the learned representations. Our proposed structural metrics measures the domain-specific details of generated composite images (fiber count, volume fraction, misalignment angle), confirming that ORDER's ordinal-aware representations contribute to diverse, high-fidelity and physically concise microstructures. 


To conclude, ORDER demonstrates that explicitly modeling property orderliness alongside cross-modal alignment is essential in composite materials, where mechanical property varies smoothly with fiber microstructure, making ordinal structure a physically grounded inductive bias rather than an architectural assumption. By bridging discrete classification paradigms with continuous regression requirements, ORDER provides a principled framework for integrating heterogeneous composite material data.  The combination of ordinal-aware representation learning and physics-based surrogates addresses core concerns around label efficiency for multimodal composite learning. As foundation models transform scientific discovery, ORDER's strategy of adapting pretrained knowledge while respecting domain-specific  structure offers viable path and insights toward more capable intelligence system for material science. Possible future improvements include developing uncertainty quantification methods for predictions and retrieval rankings, and exploring alternative strategies for handling multiple competing target properties.

\section{Methods}

\subsection{Data preparation}
\label{sec:method_data}

\begin{table}[!t]
\centering
\caption{Statistical details of the Composite dataset.}
\label{tab:composite_detail}
\vspace{-10pt}
\resizebox{0.95\linewidth}{!}{
\begin{tabular}{@{}lccc|cc@{}}
  \toprule
  & \begin{tabular}[c]{@{}c@{}} 
    Fiber \\ count \end{tabular} & MMA ($^\circ$) & Vf (\%) & \begin{tabular}[c]{@{}c@{}} Elongation \\ (mm) \end{tabular} & \begin{tabular}[c]{@{}c@{}} Yield strength \\ (MPa) \end{tabular} \\ \midrule
  Minimum  & 13.0 & 0.00 & 31 & 1.6e-4 & 13.0 \\
  Maximum  & 26.0 & 4.97 & 65 & 0.02 & 2149.8 \\
  Average  & 21.8 & 1.76 & 48 & 0.004 & 488.5 \\ \bottomrule
\end{tabular}}
\vspace{-10pt}
\end{table}

The Composite dataset is generated using the Short Fiber/Continuous Fiber RVE Generator in ANSYS Material Designer. Each RVE has dimensions of $40\times40\times2,\mu\text{m}$ and consists of an isotropic epoxy matrix reinforced with randomly distributed continuous cylindrical T700 carbon fibers, with non-overlapping geometric constraints enforced during fiber placement. T700 fibers are modeled with Young's modulus $E=230\,\text{GPa}$, tensile strength $4900\,\text{MPa}$, thermal conductivity $k=9.6\,\text{W/mK}$, and electrical conductivity $\sigma=6.25\times10^{4}\,\text{S/m}$. The epoxy matrix is modeled as an isotropic linear elastic material with Young's modulus $E=3.59\,\text{GPa}$, Poisson's ratio $\nu=0.3$, thermal conductivity $k=0.15\,\text{W/mK}$, and electrical conductivity $\sigma=0.56\,\text{S/m}$. Both strictly unidirectional and directionally perturbed fiber configurations are included to capture manufacturing-induced fiber misalignment. A conformal finite element mesh is generated per RVE to ensure accurate stress transfer and field continuity at the fiber-matrix interface. Periodic boundary conditions (PBCs) are applied for computational homogenization, from which the effective tensile yield strength and elongation are extracted alongside the corresponding cross-sectional microstructure image to form each multimodal pair.
We try to model as many possible descriptor choices as possible, while only retaining descriptors successfully modeled by the Ansys Material Designer. The resultant dataset includes 436 pairs of descriptors, their corresponding target properties and microstructures. The explored range of descriptor and target property is shown in \cref{tab:composite_detail}. The descriptors include Vf for deciding the number of fiber in the matrix, MMA that decides the rotation angle of fibers and the corresponding orientation tensor. The target properties include tensile yield strength and elongation.

For the obtained microstructures, we center-crop and resize them into size (224,224) to fit the input resolution of mainstream vision encoders. Random horizontal and vertical flip of probability 0.1 are applied.  Since the colors of original images are for display purpose only, we further transform all composite microstructures into grayscale images. All tabular descriptors are unchanged, and treated as continuous input in the table encoder.

We follow the data and preprocessing pipeline in MatMCL~\cite{sgpt} for the Nanofiber dataset. The continuous descriptors are standardized using z-score normalization, while discrete descriptors are unchanged.  

\subsection{Ordinal-aware image-tabular pretraining}
\label{sec:method_pretrain}
The pretraining of ORDER is performed on image and tabular inputs to achieve cross-modal alignment and in-modality orderliness. Recall the notations in \cref{sec:order}. Given original image inputs $v_i$ and tabular inputs $t_i$, $i \in [1,N]$ where $N$ is the number of sample pairs, the encoders $E_t,E_v$ map them to a shared latent space:
\begin{equation}
  h_i^t = E_t(t_i), h_i^v = E_v(v_i),
\label{eq_def}
\end{equation}
where $h \in \mathbb{R}^{d}$, $d=128$ is feature dimension.
We propose to optimize the following losses based on the two features.

\subsubsection{Cross-modal contrastive loss}
The cross-modal contrastive loss was first adopted by Zhang \textit{et al.} \cite{zhang2022contrastive} to align across modalities, and further proved effective on large scale data by Radford \textit{et al.} \cite{radford2021learning}. The loss encourages higher similarity between matched cross-modal pairs (positive pairs) and low similarity between other pairs (negative pairs):
\begin{equation}
  \mathcal{L}_{v \to t} = - \sum_i \log \frac{\exp(h_i^v h_i^t / \tau)}{\sum_{j \neq i} \exp(h_i^v h_j^t / \tau) + \exp(h_i^v h_j^v / \tau)},
\label{eq_clip}
\end{equation}
where $\tau$ is temperature parameter and set to 0.1 in this work. We can define $\mathcal{L}_{t \to v}$ similarly. The overall alignment loss is defined as:
\begin{equation}
  \mathcal{L}_{\mathrm{align}} =  (\mathcal{L}_{v \to t} + \mathcal{L}_{t \to v}) / 2.
\label{eq_align}
\end{equation}
Note that the alignment loss $\mathcal{L}_{\mathrm{align}}$ has been widely adopted for matching various modalities \cite{sgpt,hager2023best,xu2021videoclip}. Optimization of the loss has become the foundation of multimodal systems.

\subsubsection{Ordinal-aware contrastive loss}
Inspired by Zha \textit{et al.} \cite{rnc}, we adopt an ordinal-aware contrastive loss on both the image and tabular modality to ensure in-modal feature orderliness with respect to their target properties. The loss encourages higher similarity between feature pairs with more similar target properties, and vice versa. The ordinal-aware contrastive loss for vision features is defined as:
\begin{equation}
  \mathcal{L}_{v} = - \frac{1}{N(N-1)} \sum_i \sum_{j \neq i}  \log \frac{\exp(h_i^v h_j^v / \tau)}{\sum_{k \in N, d(i,k) \ge d(i,j)} \exp(h_i^v h_k^v / \tau)},
\label{eq_sing_rnc}
\end{equation}
where $d(i,j)$ computes the target property distance between the $i_{\rm th}$ and $j_{\rm th}$ sample. This work uses L2 distance between normalized target property values for the computation of $d(\cdot)$. The loss $\mathcal{L}_{t}$ for tabular features can be defined similarly with \cref{eq_sing_rnc}, and the overall ordinal loss is defined as:
\begin{equation}
  \mathcal{L}_{\mathrm{order}} = \mathcal{L}_{v} + \mathcal{L}_{t}.
\label{eq_rnc}
\end{equation}
The optimization of \cref{eq_rnc} encourages the feature space to imitate the distribution of their corresponding target space. While \cref{eq_rnc} requires computing target property distances, it does not directly access the target property values. Instead, only the relative order between sample pairs is needed to determine the negative samples. 

\subsection{Distance-based contrastive baselines}
\label{sec:method_baselines}
To benchmark ORDER against methods that also encode continuous property information, we implement two distance-based contrastive baselines, both sharing the same model architecture  and cross-modal alignment loss $\mathcal{L}_\text{align}$ as ORDER, differing only in their within-modality property-aware component.

\subsubsection{Distance-Weighted Contrastive Loss}
Distance-Weighted Contrastive Loss (DWCL) is a soft-weighting baseline we construct, inspired by the supervised contrastive loss~\cite{khosla2020supervised} and extending its binary positive membership to continuous property-similarity weights. DWCL replaces the hard binary positive of standard contrastive learning with soft, property-similarity-proportional weights over all pairs.
The adaptation addresses the mismatch between SupCon and the materials setting. SupCon relies on discrete class labels to define positives, but composite properties such as yield strength and elongation are continuous quantities with no natural class boundaries. Therefore, we introduce  soft weight $w_{ij}$ so that the contrastive signal can vary smoothly with property proximity. Property vectors are z-score normalized before computing $w_{ij}$. Given L2-normalized embeddings $\mathbf{z}_i$ and z-score-normalized property vectors $\mathbf{y}_i$, pairwise soft weights are:
\begin{equation}
  w_{ij} = \frac{\exp(-\|\mathbf{y}_i - \mathbf{y}_j\|_1 / \sigma)}{\sum_{k \neq i} \exp(-\|\mathbf{y}_i - \mathbf{y}_k\|_1 / \sigma)}, \quad j \neq i,
\end{equation}
where bandwidth $\sigma=1.0$ controls the decay rate (in z-score label space). The DWCL objective is:
\begin{equation}
  \mathcal{L}_\text{DWCL} = -\frac{1}{N} \sum_{i=1}^{N} \sum_{j \neq i} w_{ij} \log \frac{\exp(\mathbf{z}_i \cdot \mathbf{z}_j / \tau)}{\sum_{k \neq i} \exp(\mathbf{z}_i \cdot \mathbf{z}_k / \tau)}.
\end{equation}

\subsubsection{Property-Gap Triplet Loss}
Property-Gap Triplet Loss (Triplet) extends the classical triplet loss~\cite{schroff2015facenet} by setting the margin proportional to the property gap between anchor-positive and anchor-negative pairs.
FaceNet's original formulation uses a fixed scalar margin $m$ applied uniformly to all triplets, regardless of how far apart the anchor-positive and anchor-negative samples are in label space. Our modification makes the margin adaptive: it scales linearly with $d_{ik} - d_{ij}$, the property gap between the negative and the positive, so that the embedding space is required to reflect the magnitude of property differences, not merely their direction.
Let $S_{ij} = \mathbf{z}_i \cdot \mathbf{z}_j$ denote the cosine similarity and $d_{ij} = \|\mathbf{y}_i - \mathbf{y}_j\|_1$ the L1 property distance. A valid triplet $(i, j, k)$ satisfies $d_{ij} < d_{ik}$, i.e., sample $j$ is the property-space positive and $k$ is the negative. The loss is:
\begin{equation}
  \mathcal{L}_\text{Triplet} = \frac{1}{|\mathcal{T}|} \sum_{(i,j,k) \in \mathcal{T}} \max\!\bigl(0,\; S_{ik} - S_{ij} + \beta(d_{ik} - d_{ij})\bigr),
\end{equation}
where $\beta=0.2$ is a global scale factor and $\mathcal{T}$ is the set of all valid triplets within the batch. The property-gap-scaled margin encourages the embedding separation to grow linearly with the property gap.

\subsubsection{Combined training objective}
Both baselines apply the within-modality loss independently to tabular and image embeddings (obtaining $\mathcal{L}_{\text{DWCL}}^{\text{tab}}, \mathcal{L}_{\text{DWCL}}^{\text{img}}, \mathcal{L}_{\text{Triplet}}^{\text{tab}}, \mathcal{L}_{\text{Triplet}}^{\text{img}}$). Combined with the cross-modal alignment loss we have the overall training objective:
\begin{equation}
  \mathcal{L}_\text{total} = \alpha \cdot \bigl(\mathcal{L}^\text{tab} + \mathcal{L}^\text{img}\bigr) + (1-\alpha) \cdot \mathcal{L}_\text{align},
  \label{eq_baseline_combined}
\end{equation}
where $\alpha$ is selected by grid search over the same candidate values used for ORDER-$\alpha$ (see \cref{sec:method_overall}).
The design of applying the property-aware loss within each modality independently ensures that each encoder produces a unimodal embedding space that is both internally ordered by property value and cross-modally aligned with its counterpart, mirroring the structure of ORDER's $\mathcal{L}_\text{order}$. This formulation creates a controlled comparison that isolates the effect of the property-encoding mechanism: soft exponential weighting (DWCL), gap-proportional margin (Triplet), or rank-based ordinal structure (ORDER). Therefore, the observed performance difference between the baselines and ORDER can be attributed to the choice of within-modality property-aware objective.

\subsection{Physics-based ordinal surrogates}
\label{sec:method_surrogates}
The ordinal-aware contrastive loss $\mathcal{L}_\text{order}$ (\cref{eq_sing_rnc}) requires only pairwise property rankings, not absolute property values. Any surrogate that preserves inter-sample ordering can therefore replace ground-truth labels, enabling pretraining with reduced annotation. We derive such surrogates from constitutive material models.

\subsubsection{Surrogate quality criterion}
We evaluate surrogates using \textbf{triple rank accuracy}: for every anchor $i$ and pair of other samples $(j, k)$, the fraction of triplets where the surrogate and the ground-truth agree on the relative ordering of pairwise distances to $i$. This mirrors the comparison mechanism in $\mathcal{L}_\text{order}$ and is more sensitive to distance-compression artifacts than pairwise rank accuracy.

\subsubsection{Composite dataset surrogate}
For CF-T700 composites, the Krenchel rule of mixtures \cite{krenchel1964fibre} gives $\sigma_c \approx V_f \cdot \sigma_f \cdot \cos^2(\mathrm{MMA}) + (1-V_f)\cdot\sigma_m$. Since all MMA values fall below $5^\circ$ (monotonically decreasing regime of $\cos^2$), we use $-\mathrm{MMA}$ as an order-preserving proxy:
\begin{equation}
  S_\text{comp} = -\mathrm{MMA} + \varepsilon \cdot V_{f,\text{norm}},
\end{equation}
where $\varepsilon = 2.44 \times 10^{-6}$ is a hierarchical tiebreaker for the 22.5\% of samples with MMA\,$=0$ (for which fiber spatial randomness, not MMA, drives property variation). The tiebreaker weight is chosen small enough not to reorder non-tied samples. This surrogate achieves 87.9\% pairwise and 77.3\% triple rank accuracy for yield strength.

\subsubsection{Nanofiber dataset surrogate}
The stiffness of Electrospun nanofiber is governed by two independent structural factors. First, fiber volume fraction $\Phi_f$ determines the total amount of load-bearing material, such that higher $\Phi_f$ increases stiffness. Second, fiber orientation relative to the test direction controls how efficiently fibers resist deformation, captured by the Krenchel orientation efficiency factor~\cite{krenchel1964fibre} $\eta_o = \langle\cos^4\varphi\rangle$ ($\varphi$ = angle between each fiber and the loading direction). The surrogate  separates the density and orientation contributions additively:
\begin{equation}
  S_\text{fib} = w_{f,\text{norm}} + \tilde{r} \cdot (2 \cdot \mathrm{dir} - 1),
\end{equation}
where $w_{f,\text{norm}}$ is the normalized weight-to-flow-rate ratio (proxy for $\Phi_f$), $\tilde{r} = r/1600$ is the normalized drum rotation speed (proxy for $\eta_o$, since faster rotation aligns fibers), and $\mathrm{dir}\in\{0,1\}$ is the test direction. The factor $(2\cdot\mathrm{dir}-1)$ takes value $+1$ for the longitudinal direction (alignment boosts stiffness) and $-1$ for the transverse direction (alignment reduces stiffness). For elongation, which is negatively correlated with stiffness, the surrogate is $-S_\text{fib}$. This achieves 76.6\% pairwise and 64.2\% triple rank accuracy for elastic modulus, substantially above the ${\sim}50\%$ random baseline.

\subsection{Model optimization}
\label{sec:method_opt}
\subsubsection{Preference-guided multi-objective optimization}
\label{sec:method_pareto}
Our ORDER simultaneously optimizes $\mathcal{L}_{\mathrm{align}}$ and $\mathcal{L}_{\mathrm{order}}$ to obtain an aligned and ordered multimodal feature space. As introduced in \cref{sec:method_pretrain}, $\mathcal{L}_{\mathrm{align}}$ treats feature pairs from the same modality as negative pairs, while $\mathcal{L}_{\mathrm{order}}$ might push in-modal features with similar target properties closer. Therefore, it is crucial to appropriately handle the optimization of these two conflicting objectives so that the effects of both optimization terms are preserved. We weight the two losses with a hyperparameter $\alpha \in (0,1)$:
\begin{equation}
  \mathcal{L} = \alpha \cdot  \mathcal{L}_{\mathrm{order}} + (1-\alpha) \cdot \mathcal{L}_{\mathrm{align}}.
\label{eq_alpha}
\end{equation}
A larger $\alpha$ encourages the model to focus more on in-modal orderliness, while a smaller $\alpha$ leads to more closely aligned cross-modal pairs. One strategy, termed as ORDER-$\alpha$ in this work, selects $\alpha$ with grid-search for each task. ORDER-$\alpha$ achieves better performance, but is time-consuming and might not be realistic in label-scarce scenarios. 

To address this, we further propose ORDER-dyn with dynamically adjusted $\alpha$ during optimization. Inspired by Mahapatra \textit{et al.} \cite{mahapatra2020multi}, we adopt a preference-guided strategy to achieve Pareto optimal weighting solutions. Specifically, we formulate the joint optimization of $\mathcal{L}_{\mathrm{align}}$ and $\mathcal{L}_{\mathrm{order}}$ as a multi-objective optimization problem, where we seek parameters that achieve the optimal trade-off between cross-modal alignment and in-modal orderliness.

For simplicity, let $\mathcal{L}_1 = \mathcal{L}_{\mathrm{order}}$ and $\mathcal{L}_2 = \mathcal{L}_{\mathrm{align}}$ denote our two training objectives, with corresponding gradients $g_1 = \nabla_\theta \mathcal{L}_1$ and $g_2 = \nabla_\theta \mathcal{L}_2$ with respect to model parameters $\theta$. Rather than using a fixed scalar weighting as in \cref{eq_alpha}, we compute an adaptive update direction $h$ at each training iteration that balances both objectives guided by validation performance. Following preference-guided multi-objective optimization \cite{mahapatra2020multi}, we model the update direction as a convex combination of training gradients: $h = G\beta$, where $\beta = [\beta_1, \beta_2]$ is weight vector with $\beta_1 + \beta_2 = 1$ and $G = [g_1, g_2] \in \mathbb{R}^{n \times 2}$ is the gradient matrix.

To dynamically guide the optimization toward solutions that generalize well to the target domain, we leverage the gradient of $\mathcal{L}_{\mathrm{align}}$ evaluated on a held-out validation set. Specifically, we compute the validation gradient $\hat{g}_v = \nabla_\theta \mathcal{L}_{\mathrm{align}}^{\mathrm{val}}$ and use it to steer the optimization direction. The validation set is randomly sampled (15\% of the original data) and remains fixed throughout training. We evaluate performance on held-out data because it suggests generalization capability and provides more stable guidance.

The optimal combination $\beta^* = [\beta_1^*, \beta_2^*]^T$ is obtained by solving a linear program that adapts based on the validation loss magnitude.  When the validation loss $\mathcal{L}_{\mathrm{align}}^{\mathrm{val}} > \epsilon$ (where $\epsilon$ is a small threshold), we maximize alignment between the update direction and the validation gradient:
\begin{equation}
\begin{aligned}
  \beta^* = \arg\max_{\beta \in S_2} \quad & (G\beta)^T \hat{g}_v \\
  \text{s.t.} \quad & (G\beta)^T g_j \geq \mathbbm{1}_{J \neq \emptyset} \cdot \hat{g}_v^T g_j, \quad \forall j \in \bar{J} \setminus J^*,\\
  & (G\beta)^T g_j \geq 0, \quad \forall j \in J^*,
\end{aligned}
\label{eq_lp_bal}
\end{equation}
where $J = \{j \mid \hat{g}_v^T g_j > 0\}$ identifies objectives whose gradients align with the validation gradient, $\bar{J} = \{1, 2\} \setminus J$, and $J^* = \{j \mid \hat{g}_v^T g_j = \max_{j'} \hat{g}_v^T g_{j'}\}$ identifies the objective most aligned with validation performance. When $\mathcal{L}_{\mathrm{align}}^{\mathrm{val}} \leq \epsilon$, we maximize the sum of gradient projections:
\begin{equation}
\begin{aligned}
  \beta^* = \arg\max_{\beta} \quad & \sum_{j=1}^{2} (G\beta)^T \cdot g_j  \\
  \text{s.t.} \quad & (G\beta)^T \cdot g_j  \geq 0, \quad \forall j \in \{1, 2\}.
\end{aligned}
\label{eq_lp_desc}
\end{equation}

This formulation ensures that when validation performance is poor ($\mathcal{L}_{\mathrm{align}}^{\mathrm{val}} > \epsilon$), the optimization prioritizes directions that improve validation alignment, potentially allowing controlled ascent on certain training objectives to escape suboptimal regions. Conversely, when validation performance is satisfactory, all training objectives are simultaneously minimized.
The indicator function $\mathbbm{1}_{J \neq \emptyset}$ equals 1 when at least one training gradient aligns with the validation gradient, and 0 otherwise, preventing unbounded ascent when no training objective benefits validation performance. The parameter update is then $\theta^{t+1} = \theta^t - \eta G\beta^*$, where $\eta$ is the learning rate. This adaptive strategy allows ORDER-dyn to automatically balance the two objectives without manual hyperparameter tuning, dynamically adjusting the trade-off based on validation feedback to achieve superior generalization performance.

\subsubsection{Low-rank adaptation for material-specific knowledge injection}
\label{sec:method_lora}
To effectively leverage pretrained knowledge from CLIP while adapting to domain-specific composite materials, we employ Low-Rank Adaptation (LoRA) \cite{hu2022lora} for parameter-efficient fine-tuning (PEFT) of the vision encoder $E_v$. LoRA introduces trainable low-rank decomposition matrices into the attention layers of the Vision Transformer, while keeping the original pretrained weights frozen.

The adoption of LoRA addresses two critical challenges in multimodal pretraining for material science. First, paired multimodal material datasets are scarce (typically with merely hundreds of data points) due to expensive characterization procedures. Full fine-tuning of pretrained models like CLIP with these data would lead to severe overfitting and destruction of pre-aligned multimodal structures \cite{gao2024clip,li2025unified}. LoRA reduces trainable parameters to less than 1\% of the original model to substantially mitigate overfitting risks. Second, LoRA preserves the robust visual representations learned from large-scale pretraining on 400M image-text pairs. This enables ORDER to benefit from both general visual understanding and material-specific features without catastrophic forgetting. Our empirical results demonstrate that LoRA-based fine-tuning consistently outperforms training from scratch or using smaller architectures like ResNet50, validating its effectiveness for multimodal material science applications.

For a pretrained weight matrix $W_0 \in \mathbb{R}^{d \times k}$ in the attention module, LoRA represents the weight update as:
\begin{equation}
  W = W_0 + \Delta W = W_0 + BA,
\label{eq_lora_upd}
\end{equation}
where $B \in \mathbb{R}^{d \times r}$ and $A \in \mathbb{R}^{r \times k}$ are trainable low-rank matrices with rank $r \ll \min(d, k)$. With learned matrices $A,B$, the forward propagation for image input $x_i$ becomes:
\begin{equation}
h_i^v = W_0x_i + BAx_i.
\label{eq_lora_forward}
\end{equation}
Matrix $A$ is initialized with random Gaussian values, while $B$ is initialized to zero, ensuring $\Delta W = 0$ at the start of training. We apply LoRA to the query, value, key and output projection matrices in all attention layers of the CLIP vision backbone with rank $r = 8$ and scaling factor $\alpha_\text{LoRA} = 16$.

\subsubsection{Overall training process}
\label{sec:method_overall}
We optimize \cref{eq_alpha} with $\alpha$ either pre-defined or decided with multi-objective framework described in \cref{sec:method_pareto}. The parameters in table encoder $E_t$ are fully fine-tuned. For thorough comparison, we provide results with both ImageNet-pretrained ViT-B/16 and CLIP-pretrained ViT-B/16 as vision encoder. The ImageNet-pretrained backbone is fully tunable and the CLIP one is LoRA-tuned to preserve its fragile multimodal details.
On all tasks, we pretrain with ORDER, CMCL, DWCL, and Triplet for 200 epochs under initial learning rate 3e-4 using Adam optimizer \cite{kingma2014adam}. Batch size is 32 for all methods and tasks. All experiments are implemented with PyTorch \cite{paszke2019pytorch} and conducted on NVIDIA L40 GPUs. 

\subsection{Downstream tasks}
\label{sec:method_down}
We construct a five-way split for each dataset to ensure strict separation between pretraining and prediction evaluation. Approximately 15\% of the full data is withheld as a held-out test set before any model training. Of the remaining 85\%, roughly 58\% is allocated to the pretraining subset ($\approx$50\% training and $\approx$8\% validation) and roughly 27\% to the prediction subset ($\approx$20\% training and $\approx$7\% validation). 
Pretraining models are trained exclusively on the pretraining training split. ORDER-dyn additionally uses the pretraining validation split for adaptive gradient weight computation. Cross-modal retrieval and microstructure generation tasks use the train set for pretraining and are evaluated on the held-out test split. For property prediction, the MLP predictor is trained on the prediction training split and evaluated on the held-out test set. This strict disjointness ensures that no test sample's image or tabular descriptor was observed during pretraining in any form, eliminating data leakage through ordinal exposure or feature memorization. Only the prediction tasks are supervised by target property values. Note that ORDER-$\alpha$ and ORDER-dyn additionally consume ground-truth property rankings during pretraining, which CMCL and MatMCL do not. ORDER-$\alpha$-surr and ORDER-dyn-surr replace these rankings with physics-derived surrogates and therefore operate under equivalent supervision to the baselines.

We also construct OOD splits for evaluation in \cref{fig_analysis}. For the Composite dataset, the test set comprises 65 samples from the high-Vf regime (Vf $\geq$ 0.501). For the Fiber dataset, the test set consists 70 samples with $f = 0.6$ (highest fiber packing density). In both datasets 37.5\% of the OOD range is retained in training to provide partial exposure.

\subsubsection{Cross-modal retrieval}
\label{sec:method_retrieval}
After pretraining, the frozen encoders $E_v$ and $E_t$ produce aligned and ordinal-aware features that enable direct cross-modal retrieval without additional training. 
For image-to-tabular retrieval, given a query image $x_q$ with extracted feature $h^v_q = E_v(x_q)$, we compute similarities with all tabular features of candidate data split:
\begin{equation}
  s_{i} = \frac{h^v_q \cdot h^t_i}{\|h^v_q\| \|h^t_i\|}, \quad i \in [1, N],
\label{eq_retrieval}
\end{equation}
where $h^t_i = E_t(t_i)$ are the precomputed tabular features. The top-$k$ candidates are selected as:
\begin{equation}
  \mathcal{R}_k = \{i_1, i_2, \ldots, i_k\} \text{ where } s_{i_1} \geq s_{i_2} \geq \cdots \geq s_{i_k}.
\end{equation}
Tabular-to-image retrieval process can be defined similarly. For all methods with hyperparameter $\alpha$ (ORDER-$\alpha$ variants and DWCL, Triplet), we set $\alpha=0.5$ for the retrieval task. The ordinal-awareness ensures that retrieved candidates not only match semantically but also exhibit similar target properties, making the retrieval results practically useful for inverse material design.

\subsubsection{Target property prediction}
\label{sec:method_predict}
We evaluate the quality of learned representations through supervised property prediction tasks. After pretraining, both encoders are frozen and used as feature extractors. A two-layer MLP predictor is trained on the extracted features to predict target properties $y_i$.
For single-modality prediction, the MLP takes either tabular features $h^t_i$ or image features $h^v_i$ as input:
\begin{equation}
  \hat{y}_i = f_{\text{MLP}}(h^t_i) \quad \text{or} \quad \hat{y} = f_{\text{MLP}}(h^v_i),
  \label{eq_predict}
\end{equation}
where $f_{\text{MLP}}$ consists of 2 hidden layers of shape $(d,d)$ with ReLU activation, and a final layer of shape $(d,1)$ that produces prediction results. Note that the target values are normalized using z-score for the training process. The optimal $\alpha$ values for all $\alpha$-tuned methods are summarized in \cref{tab:alpha_values}. We use these values in prediction tasks.

\begin{table}[!t]
\centering
\caption{Optimal $\alpha$ values for prediction tasks by method, backbone, and dataset.}
\label{tab:alpha_values}
\vspace{-4pt}
\resizebox{0.95\linewidth}{!}{
\begin{tabular}{llcc}
\toprule
Method & Backbone & Nanofiber & Composite \\
\midrule
ORDER-$\alpha$      & CLIP      & 0.9 & 0.2 \\
ORDER-$\alpha$      & ImageNet  & 0.9 & 0.2 \\
ORDER-$\alpha$-surr & CLIP      & 0.9 & 0.2 \\
ORDER-$\alpha$-surr & ImageNet  & 0.5 & 0.9 \\
DWCL                & CLIP/ImageNet & 0.9 & 0.9 \\
Triplet             & CLIP/ImageNet & 0.9 & 0.9 \\
\bottomrule
\end{tabular}}
\vspace{-4pt}
\end{table}

For multimodal fusion prediction, we first concatenate features from both modalities and project them to a unified dimension through an additional fusion layer:
\begin{equation}
  h^{\text{fuse}}_i = f_{\text{proj}}([h^t_i; h^v_i]) \in \mathbb{R}^d,
  \label{eq_fuse}
\end{equation}
where $[\cdot; \cdot]$ denotes concatenation and $f_{\text{proj}}$ is a linear projection layer of shape $(2d,d)$ that reduces the concatenated feature dimension back to $d$. The fused representation is then fed into the MLP predictor for prediction as in \cref{eq_predict}.
This fusion strategy enables the model to leverage complementary information from both modalities that overcomes modality-specific limitations observed in single-modality predictions.

Finally, the predictor and projection for fusion features are optimized with standard mean squared error (MSE) loss:
\begin{equation}
  \mathcal{L}_{\text{p}} = \frac{1}{N} \sum_{i=1}^N \|\hat{y}_i - y_i\|_2^2. 
\end{equation}
For all methods and tasks, we train with learning rate 5e-4 for 100 epochs using Adam optimizer and batch size 32. Early stopping is applied on Composite dataset to prevent overfitting: if the evaluation loss does not improve for 20 consecutive epochs, the training terminates and the results of the best evaluation epoch are reported.

\subsubsection{Descriptor-conditioned microstructure generation}
\label{sec:method_generation}
To demonstrate the generative capability of ORDER, we implement descriptor-conditioned microstructure generation following the framework of DALL-E 2 \cite{ramesh2022hierarchical}. The generation process consists of two stages: (1) a diffusion prior network that generates image embeddings conditioned on tabular descriptors, and (2) a diffusion decoder network that synthesizes microstructures from these generated embeddings.

Based on pre-extracted features $h_i^t$ and $h_i^v$, the prior network $P$ is designed to model the conditional distribution $p(h^v | h^t)$.  During training, we corrupt the real image feature $h^v_i$ with Gaussian noise at timestep $k$ to obtain $h^v_{i,k}$. The prior network is trained to predict the clean image feature $h^v_i$ given the noisy feature and the tabular condition:
\begin{equation}
  \mathcal{L}_{\text{prior}} = \mathbb{E}_{i, k} \| h^v_i - P(h^v_{i,k}, h^t_i, k) \|^2_2,
\label{eq_train_prior}
\end{equation}
where $P$ outputs the predicted unnoised embedding. At inference, we sample a random Gaussian noise vector and iteratively denoise it using $P$ conditioned on $h^t_i$ to obtain the predicted image feature $\hat{h}^v_i$.

The decoder network $D$ reconstructs microstructure images conditioned on these image features using a standard Denoising Diffusion Probabilistic Model (DDPM) \cite{ho2020denoising}. We employ $D$ to iteratively predict a noise component $\epsilon$ added to the image at timestep $k$ with the following loss function:
\begin{equation}
  \mathcal{L}_{\text{decoder}} = \mathbb{E}_{i, \epsilon, k} \|\epsilon - D(v_{i,k}, h^v_i, k)\|^2_2,
\label{eq_decoder}
\end{equation}
where $v_{i,k}$ is the noised image at timestep $k$ obtained via the forward diffusion process on the ground truth image $v_i$.

At inference time, given descriptors $t_0$ of interest, we first extract tabular features: $h^t_0 = E_t(t_0)$. We then generate the image feature $\hat{h}^v_0$ via the prior network's reverse diffusion process and finally synthesize the image $x^{\text{gen}}_0$ using the decoder $D$ conditioned on $\hat{h}^v_0$. The image $x^{\text{gen}}_0$ is synthesized by iteratively applying the reverse diffusion update~\cite{ho2020denoising} from $k=K$ to $k=0$, gradually recovering the clean image from noise.
For both datasets, the prior and decoder networks are trained for 200 epochs and $\sim$150k iterations, respectively, with $K=1000$ and learning rate 1e-4 using Adam optimizer.  The ordinal structure in the learned feature space encourages the generation of microstructures with properties consistent with the input specifications.

\subsubsection{Domain-specific structural generation metrics}
Beyond general image-quality metrics, we evaluate the physical fidelity of generated Composite microstructures using three domain-specific descriptors extracted by classical computer vision directly from generated images. All detectors are validated against ground-truth tabular labels on real composite training images.

\textbf{Fiber count} (det\_N, proxy for NumFibers). Each fiber appears as a roughly circular or oval-shaped cross-section. The detector tiles each grayscale image into a 3$\times$3 grid, applies Gaussian blur ($5\!\times\!5$ kernel, $\sigma$ auto) and Canny edge detection (thresholds 20/60), finds all closed boundaries in the grid, retains those with area in $(0.1\%, 25\%)$ of one tile, and counts only those whose centroids lie in the central tile. Because Composite images are produced with periodic boundary conditions, fibers crossing an image edge continue to the opposite edge and appear as incomplete boundaries in the original frame. The 3$\times$3 tiling completes these fibers before counting. A systematic under-count factor of $\approx 0.66\times$ arises from occasional merging of adjacent boundary fibers, which does not affect correlation-based evaluation. The detector achieves strong validated Spearman correlation with ground-truth fiber count.

\textbf{Gradient circular variance} (cc\_var, inverse proxy for MMA). The mean misalignment angle MMA controls fiber tilt. When MMA $= 0$, all fiber axes are perpendicular to the image plane and the cross-sections are circular, so edge gradients point uniformly in all directions. When MMA $> 0$, tilted fibers yield oval cross-sections, and gradients concentrate perpendicular to the long axis of each oval rather than spreading uniformly. cc\_var measures how uniformly gradient orientations are spread:
\begin{equation}
  \mathrm{cc\_var} = 1 - \left|\frac{1}{N}\sum_{k=1}^{N} e^{2i\theta_k}\right|,
  \label{eq_circ_var}
\end{equation}
where $\theta_k$ are the edge gradient orientation angles detected across the image and $N$ is their count. The quantity inside the absolute value measures how aligned these orientations are. When orientations concentrate in one direction the magnitude approaches 1, whereas uniformly spread orientations produce a magnitude near 0. Subtracting from 1 yields a variance-like measure: cc\_var $= 1$ when orientations are uniformly distributed (circular cross-sections, low MMA) and cc\_var $\to 0$ when they concentrate in one direction (oval cross-sections, high MMA). The metric therefore achieves negative validated Spearman correlation with MMA.

\textbf{Contour area fraction} (\textit{Vf\_est}, proxy for Vf). The fiber volume fraction is estimated geometrically from the detected fiber boundaries:
\begin{equation}
  \widehat{V}_f = \frac{\sum_k A_k}{H \times W},
  \label{eq_vfest}
\end{equation}
where $A_k$ is the enclosed area of the $k$-th detected fiber boundary, measured in the tiled image so that fibers crossing an image edge are fully captured, and $H \times W$ is the original image area. The metric achieves strong validated Spearman correlation with ground-truth fiber volume fraction.

For evaluation, all three descriptors are detected independently on the real image and on each generated image per sample. The mean over generated images gives the sample-level generated value. We report Spearman $\rho$ (whether relative ordering across samples is preserved) and mean absolute error (MAE, per-sample deviation from ground truth). For the Nanofiber dataset, the underlying images are experimental SEM micrographs with weaker and more indirect relationships between image texture and tabular descriptors. We therefore apply the standard image-quality metrics such as FID following previous work~\cite{sgpt} to the Nanofiber generation evaluation.

\section{Author contributions}
Xinyao Li designed and implemented the overall method, wrote this manuscript, and conducted experiments on the training, evaluation and analysis of the method. 
Hangwei Qian participated in the conceptualization  of the method, assisted  the writing of the manuscript, and led the collection and organization of the Composite dataset. 
Jingjing Li supervised the project and assisted  the writing of the manuscript.
Ivor Tsang led this project and participated in the Composite data construction and method conceptualization process. 
All authors reviewed and approved the final manuscript.

\section{Competing interests}
The authors declare no competing interests.

\section{Code availability}
Code is availability at \href{https://github.com/TL-UESTC/ORDER}{https://github.com/TL-UESTC/ORDER}.

\section{Data availability}

The Nanofiber dataset is publicly available at \href{https://github.com/wuyuhui-zju/MatMCL}{https://github.com/wuyuhui-zju/MatMCL}. Our Composite dataset is a part of an ongoing project that cannot be published online yet, but the data can be requested from \href{Qian\_Hangwei@a-star.edu.sg}{Qian\_Hangwei@a-star.edu.sg}. The data will also be immediately available online when the project finishes.

\ifCLASSOPTIONcaptionsoff
  \newpage
\fi



\bibliographystyle{IEEEtran}
\bibliography{main}
%

%


\end{document}